%% file: acl_latex.tex
\pdfoutput=1

\documentclass[11pt]{article}
\usepackage[table,xcdraw]{xcolor}
\usepackage[]{acl}

\usepackage{times}
\usepackage{latexsym}

\usepackage[T1]{fontenc}

\usepackage[utf8]{inputenc}

\usepackage{microtype}

\usepackage{tabularx}
\usepackage{graphics}
\usepackage{graphicx}

\usepackage{amsmath}

\usepackage{cleveref}
\usepackage{pgffor}

\newcommand\MethodName{PELMS}
\newcommand\PrimeraName{Primera}
\newcommand\PegasusXName{Pegasus-X}
\newcommand\PegasusName{Pegasus}

\newcommand\PretrainDatasetName{MultiPT}
\newcommand\RottenTomatoesName{MetaTomatoes}

\title{\textsc{\MethodName}: Pre-training for Effective Low-Shot \\Multi-Document Summarization}

\author{Joseph J. Peper, \hspace{0mm}  Wenzhao Qiu, \and Lu Wang \\
  Computer Science and Engineering \\
  University of Michigan \\
  Ann Arbor, MI \\
  \texttt{\{jpeper, qwzhao, wangluxy\}@umich.edu} \\}

\begin{document}
\maketitle 
\begin{abstract}
We investigate pre-training techniques for abstractive multi-document summarization (MDS), which is much less studied than summarizing single documents. 
Though recent work has demonstrated the effectiveness of highlighting information salience for pre-training strategy design, it struggles to generate abstractive and reflective summaries, which are critical properties for MDS. 
To this end, we present \textbf{\MethodName{}}, a pre-trained model that uses objectives based on semantic coherence heuristics and faithfulness constraints with unlabeled multi-document inputs, to promote the generation of concise, fluent, and faithful summaries. 
To support the training of \MethodName{}, we compile \textbf{\PretrainDatasetName{}}, a multi-document pre-training corpus containing over 93 million documents to form more than 3 million unlabeled topic-centric document clusters, covering diverse genres such as product reviews, news, and general knowledge.
We perform extensive evaluation of \MethodName{} in low-shot settings on a wide range of MDS datasets. Our approach consistently outperforms competitive comparisons with respect to overall informativeness, abstractiveness, coherence, and faithfulness.

\end{abstract}

\begin{figure}[t]
\centering
     \includegraphics[width=\columnwidth]{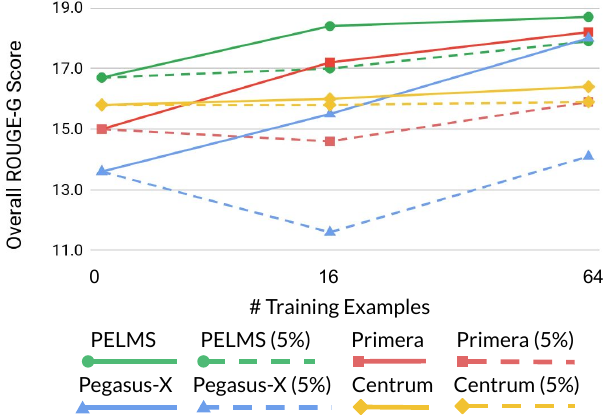}
      \caption{
      Low-shot pre-trained model performance on the multi-document summarization (MDS) task with ROUGE-G scores (geometric mean of ROUGE-1/2/L) aggregated over 6 MDS benchmarks. We compare with fully-supervised training (solid lines) and with adapter training (dashed lines) where only 5\% of model parameters are tuned. \MethodName{} achieves stronger overall performance at all data quantities and training methods when compared to competitive long-input summarization models.
      }
     \label{fig:low_shot_results}
     \vspace{-7pt}
\end{figure}

\section{Introduction}

Abstractive multi-document summarization (MDS) aims to generate concise and coherent summaries that capture key information from a set of related documents. 
Advances in transformer-based language modeling have demonstrated the benefits of summarization-oriented language model pre-training, including strategic masking and de-noising of the input \cite{zhang2019pegasus}, on a wide range of single-document summarization (SDS) tasks.
However, the direct application of these SDS models to MDS tasks often results in poor cross-document salience detection and poor information aggregation across multiple sources.

MDS-specific solutions have been developed to address limitations apparent in existing work; for example, \citet{primera} introduce \PrimeraName{}, a novel MDS-specific pre-training objective using entity frequency as an approximation for information salience. However, the quality of such pre-training objectives is often reliant on brittle topic alignment methods (such as n-gram overlap, or cross-document entity linking) which struggle to generalize to an open set of domains, yielding outputs with with subpar summary \textit{informativeness}. 

Additionally, other critical MDS properties such as multi-document \textit{faithfulness} have been overlooked by previous MDS pre-training, producing outputs that are inconsistent with their inputs. 
Similarly, summary \textit{coherence} is inadequately addressed, with most gap-sentence generation (GSG) style objectives simply denoising masked phrases in their original ordering to form the training target; these approaches may work well in conventional single-document settings, however they can introduce positional biases~\cite{deyoung2023multi_synthesis} in the multi-document setting due to the arbitrary ordering of documents within the input, impacting both \textit{coherence} and \textit{informativeness}. 

The advent of large language models (LLMs) has accelerated progress on summarization tasks including MDS, producing summaries that better address the aforementioned characteristics. Unfortunately, there are still many real-world applications for which LLM-based summarization is impractical or impossible. For example, many practical summarization applications have niche target domains (necessitating domain-specific adaptation) or require compute-efficient local inference due to privacy concerns and/or regulatory compliance demands, motivating the continued need for conventional pre-trained language models that can \textit{rapidly enable proficient MDS performance with minimal training data and/or compute.}

To address this problem, we propose \textbf{\MethodName{}}, a method of \underline{P}re-training for \underline{E}ffective \underline{L}ow-Shot \underline{M}ulti-document \underline{S}ummarization that promotes informativeness over a diverse range of text genres, ranging from consumer and editorial opinion, to news articles, and to scientific papers, while also improving abstractiveness, faithfulness, and coherence of summaries.\footnote{Code and model are made available at \url{https://github.com/jpeper/pelms}.} 
We accomplish this by utilizing efficient semantic clustering to detect topic-aware salience at the sentence level, thus avoiding the drawbacks of relying solely on lexical similarity and brittle entity extraction methods employed by previous approaches.
Our pre-training examples are constructed by selecting representative sentences from clusters that balance between topical salience and consistency to the source, for improved informativeness and faithfulness of summaries. Unlike the popular gap-sentence generation (GSG) objective, we remove input sentences instead of masking them to promote abstractiveness, and carefully select and order sentences to maintain coherence. 

\begin{figure}[t]
\centering
     \includegraphics[width=\columnwidth]{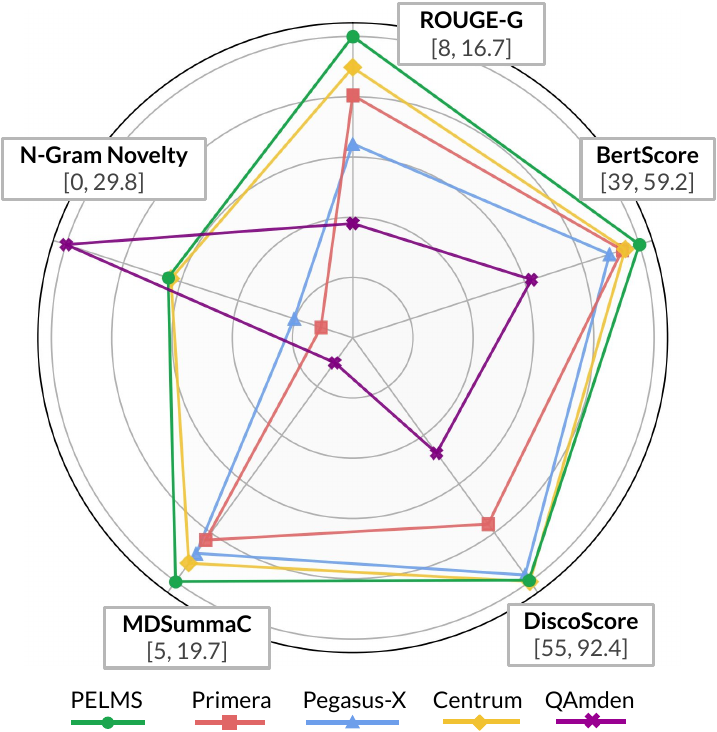}
      \caption{
      Overview of overall zero-shot performance across five metrics capturing \textbf{relevance} (ROUGE-G, BertScore), \textbf{abstractiveness} (N-gram Novelty), \textbf{coherence} (DiscoScore), and \textbf{faithfulness} (MDSummaC). For readability each metric is displayed on a unique scale, with [\textit{a}, \textit{b}] respectively indicating the minimum and maximum values for each axis. \MethodName{} achieves the best combination of performance over all key summary characteristics. 
      }
     \label{fig:radar}
     \vspace{-7pt}
\end{figure}

To support MDS pre-training, we further curate \textbf{\PretrainDatasetName{}}, a new multi-document pre-training dataset with over 3 million topic-aligned document clusters, derived from over 93 million documents---an order of magnitude larger than existing MDS pre-training corpora. 
We then conduct an extensive evaluation against various performant long-input transformer architectures on six MDS evaluation datasets spanning domains including news, science, consumer reviews, and media, including a newly introduced paragraph-length meta-summary generation dataset, \textbf{MetaTomatoes}. MetaTomatoes presents a challenging task of summarizing RottenTomatoes critics' movie reviews that necessitates strong cross-document knowledge aggregation capabilities to handle conflicting opinions. 
Our results demonstrate the strong MDS performance of \MethodName{}, particularly in low-shot settings, highlighting the alignment between our pre-training strategy and multi-document summarization. Concretely,

\begin{itemize}
    \item In the zero-shot setup, we achieve increased summary informativeness (ROUGE and BertScore) over the previous state-of-the-art pre-trained MDS models, while yielding \textit{higher or comparable faithfulness, abstractiveness, and coherence scores}. 
    In particular, we perform particularly well on review domains requiring significant cross-document synthesis. Human judges further validate that our method produces summaries with improved grammaticality, referential clarity, coherence, and faithfulness. Additionally, we incorporate length control during pre-training, and observe zero-shot gains when a desired length limit is specified.

    \item We perform experiments in supervised settings with both full-parameter and parameter-efficient training. \MethodName{} \textit{outperforms the comparisons in ROUGE and BertScore across the board}, while maintaining the best balance of faithfulness, abstractiveness, and coherence. 
    Interestingly, we observe {full-parameter training achieves higher informativeness and abstractiveness}, while {adapter-based training better maintains input faithfulness}. Our findings validate the need for thorough summarization evaluation that yields insights into model behavior beyond the standard ROUGE measurement. Notably, we find \MethodName{} \textit{can match or exceed overall GPT-3.5 and GPT-4.0 performance on informativeness, coherence, and faithfulness with as few as 16 and 64 training examples}, respectively.
    
    \item We ablate our contributions, finding both our technique and pre-training data are valuable in improving MDS performance. Concretely, we find 1) our \MethodName{}{} objective is performant with several different base long-input architectures, and 2) existing techniques benefit from pre-training on our \PretrainDatasetName{} pre-training corpus.
    
\end{itemize}

\section{Related Work}
\subsection{Multi-Document Summarization}
Multi-document summarization (MDS) poses unique challenges since their inputs are often lengthy, consisting of tens or hundreds of related articles, which necessitates computationally-efficient methods \cite{led-global-attention, pasunuru-etal-2021-efficiently, brazinskas-etal-2022-efficient}. 
More importantly, unlike single-document inputs, these multi-documents contain a mix of complementary, redundant, or sometimes contradictory information \cite{hendrickx2009reducing_redundancy, radev-2000-common}. 
Existing systems struggle with effectively aggregating knowledge from multiple documents, often overly favoring information from one document or overfitting to positional biases within the input \cite{wolhandler2022multi_how_multi_is_mds}.
We also note the lack of existing large-scale pre-training datasets for the MDS tasks, with existing multi-document modeling approaches confined to using relatively small or synthesized pre-training corpora \cite{cdlm-cross-document-language-modeling, primera, caciularu-etal-2023-peek-qamden} or extends pre-trained models designed with SDS objective.

\begin{figure*}[t]
\centering
     \includegraphics[width=\textwidth]{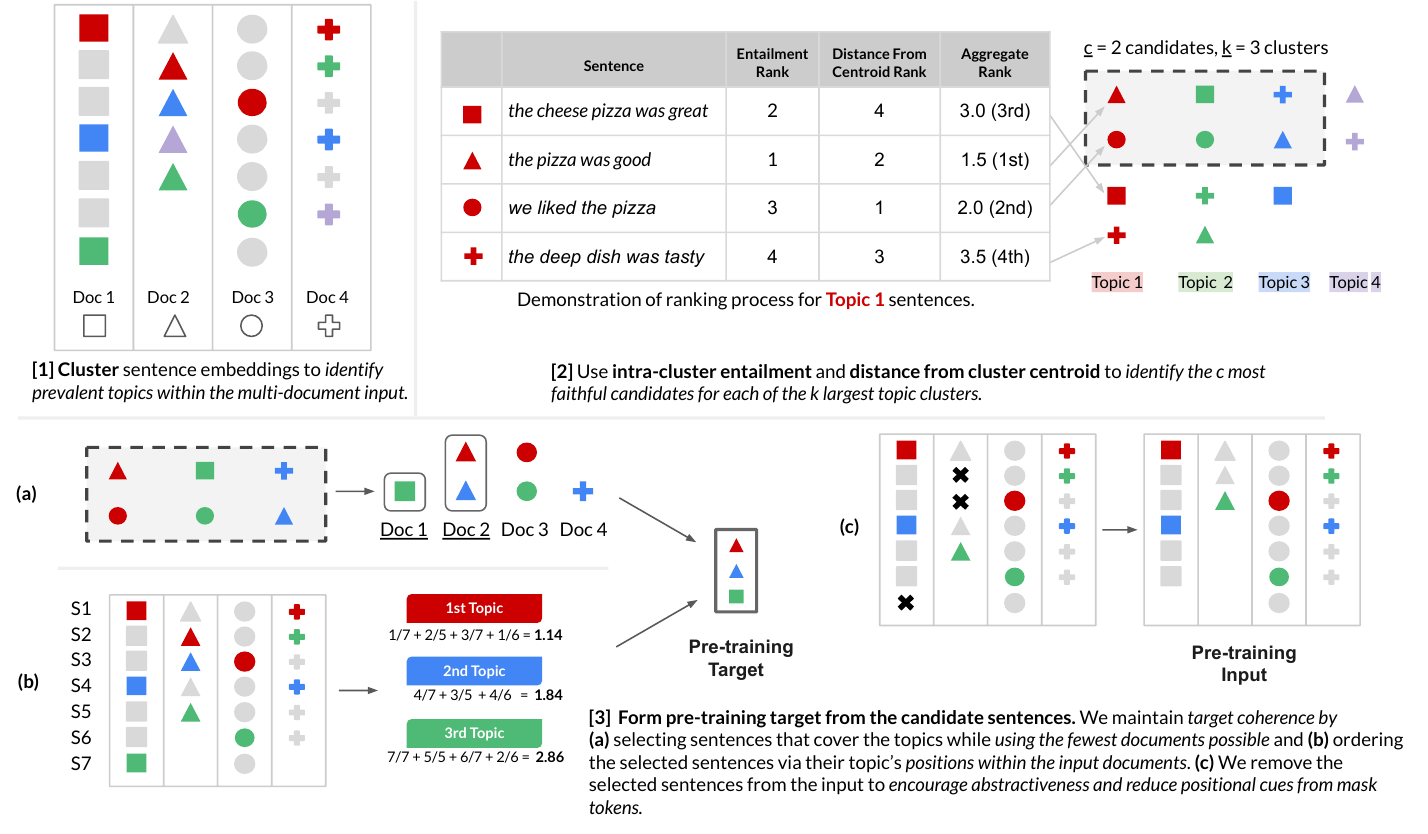}
      \caption{Overview of the \MethodName{}\ pre-training approach. The pre-training target is formed by selecting a set of representative sentences covering frequently occurring topic clusters within the input. Target sentences are intentionally selected to enhance \textbf{faithfulness} and arranged to improve target \textbf{coherence}.
      }
     \label{fig:task_overview}
     \vspace{-7pt}
\end{figure*}

\subsection{Summarization Pre-training}
Pre-trained language models (PLMs) have become the dominant paradigm in natural language processing tasks, and there has been significant work in developing pre-training strategies tailored to specific NLP subtasks, such as summarization. 
For instance, the widely used \PegasusName{} single-document summarization model \citep{zhang2019pegasus} is pre-trained using \textit{Gap Sentence Generation} (GSG), a custom denoising-style pre-training objective. In this approach, salience heuristics are used to identify and \textit{mask} the most important sentences in the input document based on lexical similarity. These masked sentences are then used as targets for the pre-training task, with the model predicting them in a sequential manner. \PegasusName{} use a ``Principal'' salience score, where sentences are selected based on ROUGE overlap with the remainder of the input. Unfortunately, this leads to repetition in the output, as redundant sentences will all score highly. 
To address this, \citet{primera} propose \PrimeraName{}, an extension of the \PegasusName{} techniques that supports multi-document inputs and mitigates the problem of redundancy through entity-based sentence grouping. They introduce the ``Entity Pyramid'' strategy for target-sentence selection, running entity extraction and grouping sentences by entity. One sentence is selected from each prevalent entity grouping, using a sentence's {average ROUGE similarity with every other input document} as the salience criteria. 
Unfortunately, the entity grouping step relies on brittle entity extraction, often failing to associate semantically-related terms and resulting in poor associations of related phrases. 
Another limitation of \PegasusName{} and \PrimeraName{} is low abstractiveness due to their pre-training masking approach: First, GSG mask tokens enable the model to learn from mask token context during denoising in pre-training, de-emphasizing the need for synthesis capabilities. Second, both techniques intentionally leave a proportion of target sentences unmasked in the input to improve model copy capacity, further downplaying abstraction skills.

\citet{puduppully-centrum} deviate from the conventional GSG approach with Centrum, a pre-training objective which selects the ROUGE-determined centroid document of each cluster as its target summary. While intuitive, this approach assumes the existence of a single reflective document, which may not hold in domains whose inputs are naturally diverse in topical content and opinion. 
\citet{caciularu-etal-2023-peek-qamden} combine cross-document information alignment and masking with question generation to perform multi-document pre-training via a question answering pre-training objective. They demonstrate strong performance in supervised settings for multi-document tasks including both question answering and summarization.

While the emphasis of these pre-training methods is on general-purpose summarization tasks (such as news summarization), similar work exists among several summarization subtasks, extending to areas such as dialog summarization \cite{li2022dionysus} and opinion summarization \cite{brazinskas-etal-2022-efficient}, with pre-training methods that address unique challenges within their tasks. In this work, we focus on general pre-training techniques for MDS that apply to a broad range of domains and applications.

In addition to general summarization pre-training to increase informativeness, methods such as FactPEGASUS \cite{wan-bansal-2022-factpegasus} attempt to guide SDS model behavior by incorporating factuality constraints during pre-training and fine-tuning. While effective, \citet{ladhak2021faithful} note that the underlying FactCC \cite{kryscinski2019evaluating_factcc} factuality model exhibits poor generalization, requiring custom domain-specific fine-tuning. Our methods generalize effectively without domain-specific design.

\begin{table*}[t]
\centering
    \small
    \include{tables/pretraining_data}
    \caption{
    Overview of the \PretrainDatasetName{} pre-training corpus. \PretrainDatasetName{} is comprised of 5 sources, each consisting of unlabeled topic-centric document clusters. Combined, they sum to over 3.6M clusters. We release \PretrainDatasetName{} along with pre-computed sentence-level auxiliary information including entities and sentence embeddings.
    }
    \label{tab:pretrain_datasets}
    \vspace{-3mm}
\end{table*}

\section {\MethodName{} Pre-training Technique}
\label{sec:technique}

The \MethodName{} pre-training strategy for multi-document summarization is outlined in Figure \ref{fig:task_overview}. Briefly, we rely on a cluster-then-select pre-training objective to generate data for training transformer models. We then follow previous work by using a Gap Sentence Generation-style objective to form pre-training targets, but instead of masking, we remove the sentences to improve abstractiveness.
Our key contributions are: 
1) improved ranking and selection of target sentence candidates to encourage summary informativeness and faithfulness, 
and 2) injecting coherence constraints during formulation of the GSG target. 
Our technique consists of the following three steps:
\begin{enumerate}
    \item \textit{Clustering sentences into topics (\S\ref{sec:technique_1_topic_detection}).} We encode and cluster the input sentences to identify prevalent topics within the input, considering the top \textit{k} topics for inclusion in the summary.
    
    \item \textit{Ranking cluster sentences by summary-worthiness (\S\ref{sec:technique_2_sentence_selection}).} We score each cluster element based on a) distance to the cluster centroid and b) entailment-based intra-cluster consistency. These rankings are used to determine which sentences are used within the pre-training target.
    
    \item \textit{Selecting and ordering target sentences to maintain summary coherence (\S\ref{sec:technique_3_target_formulation}).} Considering the \textit{c} highest-scoring examples from each topic cluster, we select target sentences (one per topic), sourcing from the fewest number of documents possible. We use this, and a topic-position ordering heuristic, to specify the output ordering.
\end{enumerate}

\subsection{Topic Detection via Sentence Clustering}
\label{sec:technique_1_topic_detection}

Information redundancy is a common phenomenon within multi-document inputs \cite{survey2022mds} and has been utilized to identify input salience, with the intuition being that topic frequency correlates with topic significance \cite{primera, nenkova2007pyramid}. Methods like \PrimeraName{} use ROUGE similarity or align sentences using entity mentions. However, these are brittle and generalize poorly, motivating the need for a more refined selection mechanism.
We propose to use \textbf{continuous semantic representations} when performing the sentence similarity comparison. Concretely, \MethodName{} embeds and clusters the input sentences, with each cluster representing a set of topic-aligned sentences.  We leverage the Sentence Transformers library \cite{reimers-2019-sentence-bert} which offers \textit{lightweight} text embeddings and a fast local-community clustering method, which are required for processing large-scale pre-training data in a tractable fashion. 
Once we have identified semantic clusters, we use this structured cluster representation to identify salient topics. Similar to the Entity Pyramid method, we use frequency as a proxy for salience. Concretely, larger clusters represent topics that are more prominent within the input, and are therefore more summary-worthy. We select from the $k$ largest clusters. See Appendix \ref{appx:pretraining_technique_details} for details on the clustering method and parameters.

\subsection {Entailment-aware Target Sentence Selection}
\label{sec:technique_2_sentence_selection}
As each top-k cluster represents a unique summary-worthy topic, we must choose a representative sentence from each cluster for inclusion within the pre-training target. \textbf{To improve the consistency of our selected sentence with the rest of the cluster}, we use a combination of cluster centrality and intra-cluster entailment to score the candidate sentences (Fig. \ref{fig:task_overview}-[2]). \PegasusName{} and \PrimeraName{} leverage simple lexical overlap to identify the most significant sentence. Similarly, we could consider simply selecting the medoid element within each topic cluster. However, methods such as \citet{wan-bansal-2022-factpegasus} find it possible to improve the faithfulness of summarization systems by imposing additional selection constraints. 
We rank the candidate sentences in two ways: 1) by distance to the cluster centroid, to capture the average semantic meaning of the cluster, and 2) by average NLI entailment with the rest of the cluster elements, as a means of maintaining input-consistent summary examples. We aggregate these two rankings using a simple unweighted Borda count \cite{borda_count} to generate a ranking of all cluster elements. We cap the NLI calculation to the 5 most central cluster elements due to the quadratic computational complexity of the intra-cluster entailment score.

\subsection{Pre-training Formulation for Improved Coherence}
\label{sec:technique_3_target_formulation}
Coherence is a key characteristic for summaries and has been extensively studied for summarization tasks including MDS \cite{christensen-etal-2013-towards_coherence, wang2016exploring_coherence}. However, GSG simply masks the highest-scoring sentences and de-noises in order of appearance within the input, which can result in incoherent outputs for arbitrarily-ordered multi-document inputs.
With our method, after identifying summary-worthy sentences, we then \textbf{select and order a set of sentences to comprise the pre-training target}. 

Considering only the $c$ highest-ranked sentences from each topic cluster, we select target sentences, one per topic, using minimum set cover to source from as few documents as possible to improve target coherence (Fig. \ref{fig:task_overview}-[3a]).

After selecting target sentences, we order them subject to the following constraints in this order of precedence: 1) sentences selected from the same document should maintain their original relative ordering, and 2) sentences should be ordered by average topic position -- e.g., `lead' topics should appear early within the target (Fig. \ref{fig:task_overview}-[3b]). Finally we remove the selected sentences from the pre-training input (Fig. \ref{fig:task_overview}-[3c]).

\section{Pre-training Details}
We pre-train a new MDS model using the \MethodName{} technique. We form pre-training examples using our new \PretrainDatasetName{} pre-training dataset as the source of document clusters. In this section we overview \PretrainDatasetName{} and the pre-training architecture used for training \MethodName{}. See Appx. \ref{appx:additional_pretraining_details} for full pre-training details.
 \vspace{-1mm}
\paragraph{\PretrainDatasetName{} Pre-training Dataset}
MDS is designed to support lengthy multi-document inputs, yet there are few available data sources for large-scale unlabeled multi-documents for pre-training.  Notably, NewsHead \citep{gu2020generating_newshead} has been used, containing clusters of news articles.
However, it is limited in magnitude for a pre-training dataset, containing only
370k document clusters.

Extending on this, we compile \PretrainDatasetName{}, a new multi-document dataset comprised of over 3 million document clusters from public data sources. It contains a wide diversity of genres (including news, general knowledge, and opinionated content), and covers a broad range of inputs with respect to document lengths and cluster sizes. \Cref{tab:pretrain_datasets} outlines the pre-training dataset.
 \vspace{-1mm}
\paragraph{Base Model}
As done in \PrimeraName{}, we use the popular sparse-attention long-input Longformer Encoder Decoder (LED) \cite{longformer_led} as our base architecture, initializing from the 464M LED-Large base model. We use the default configuration, and follow \PrimeraName{} in inserting a global \texttt{<doc-sep>} token after each input document as this enables improved cross-document communication \cite{cdlm-cross-document-language-modeling}.

\begin{table*}[t]
    \centering
    \small
    \include{tables/_full_zero_shot_analysis}
    \caption{
    Results on zero-shot multi-document summarization. 
    Values in \colorbox[HTML]{B1CBF2}{blue} indicate where \MethodName{} outperforms all baselines for a given dataset. 
    \colorbox[HTML]{B7E1CD}{Green} indicates the best performance when averaged over all six datasets. 
    We see \MethodName{} is able to achieve strong informativeness (ROUGE, BertScore) and coherence (DiscoScore) while achieving the best combination of faithfulness (MDSummaC) and abstractiveness (N-gram Novelty).
    }

    \label{tab:full_zero_shot_results}
\end{table*}

\section{Evaluation Setup}

\subsection{Evaluation Datasets}

We use five existing MDS datasets spanning news, opinion, and scientific domains in our evaluation, and additionally curate \textbf{\RottenTomatoesName{}}, a meta-summary generation dataset for critics' film reviews. Table \ref{tab:eval_datasets} overviews the evaluation dataset statistics. Appendix \ref{appx: datasets_extra} provides further dataset details. 

\subsection{Evaluation Metrics}
Previous analysis of MDS techniques has largely focused on ROUGE evaluation, supplemented by expensive human evaluation. In this work, we are the first to systematically explore the behavior of pre-trained MDS models across a wide variety of summary evaluation metrics. We evaluate summary \textit{informativeness} with \textbf{ROUGE} and \textbf{BertS}core, \textit{coherence} with \textbf{DiscoS}core, \textit{faithfulness} with our new \textbf{MDSummaC} metric, and \textit{abstractiveness} with \textbf{N-gram Novelty}. Full evaluation metrics details are covered in Appendix \ref{appx:eval_metrics}.

\begin{table*}[t]
    \centering
    \small
    \include{tables/_overall_summary}
    \caption{
    Overall summary of performance across zero-shot and supervised splits. 
    For the supervised splits, we perform adapter-based fine-tuning (updating only 5\% of model weights), and full-parameter fine-tuning. 
    \colorbox[HTML]{B1CBF2}{Blue} indicates our model outperforms all baselines for a specific tuning method (i.e., when comparing just full-parameter or just adapter results).  Values in \colorbox[HTML]{B7E1CD}{green} indicate our model achieves top performance \textit{overall} for its data quantity split. Results are averaged over all 6 datasets. We see most metrics improve with more data, although often at the expense of faithfulness. Other than n-gram novelty, \MethodName{} with 16 training examples outperforms GPT-3.5-Long, and with 64 examples outperforms GPT-4.0.
    }
    \label{tab:full_main_results}
\end{table*}

\subsection{Baseline Models}
\label{sec:baseline_models}
We compare our \MethodName{} technique against four performant long-input pre-trained summarization models to understand their behavior in both zero-shot and supervised settings:

\textbf{\PrimeraName{}} \cite{primera} is a 464M model initialized from LED-large, and trained on NewSHead \cite{gu2020generating_newshead} with the previously described entity salience pre-training objective. It has achieved strong performance in multi-document summarization among existing pre-trained language models, and serves as a strong baseline. \\

\textbf{\PegasusXName{} }\cite{phang2022pegasusx} is a 568M model with a long-input extension of the popular Pegasus model \cite{zhang2019pegasus}. It uses block-staggered local attention to enable efficient scaling to long inputs. \PegasusXName{} has demonstrated strong performance on long-input tasks, outperforming other similar-sized long-input models such as LongT5 \cite{guo2022longt5} and LED. \\

\textbf{Centrum} \cite{caciularu-etal-2023-peek-qamden} is a 464M model initialized from LED-large and trained on NewSHead, with the same sparse attention setup as \PrimeraName{}. It calculates a centroid document using ROUGE similarity and performs ``leave-one-out'' training, using this document as the target summary. \\

\textbf{QAMDen} \cite{puduppully-centrum} is a 464M model also initialized from LED-large using the Primera-style sparse attention. The pre-training task consists of multi-document question answering, using silver-labeled question-context-answer tuples that are automatically derived from unlabeled data using ROUGE salience calculation and question generation models. The training source is NewSHead document clusters, using the silver-labeling process to create 4.3M unique QA training instances.

\subsection{Supervised Fine-tuning Methods}
In our supervised experiments, we train the model using labeled summarization data. We perform standard training, updating all model parameters during training. Results are averaged over 5 runs with unique random seeds. Additionally, we explore using parameter-efficient fine-tuning (PEFT) to understand whether the evaluated methods are capable of converging when updating fewer model parameters; for the parameter-efficient method, we use the popular Adapter \cite{pmlr-v97-houlsby19a_adapters} PEFT method which freezes the original model weights and trains only a small percentage of parameters.

\section{Results}
We perform a rigorous comparison of \MethodName{} and competitive baseline models in both \textbf{zero-shot} and \textbf{supervised} settings (using both full parameter tuning and parameter-efficient fine-tuning). 
Tables \ref{tab:full_zero_shot_results} and \ref{tab:full_main_results} list the overall results in both zero-shot and supervised settings. Appendix \ref{sec:experiment_details_appendix} outlines the full details of our zero-shot and supervised evaluation configuration.

\subsection{Zero-Shot Evaluation}
Table \ref{tab:full_zero_shot_results} outlines the zero-shot results. We analyze the behavior of our method in the zero-shot case as this setting directly reflects alignment between our proposed pre-training objective and the downstream MDS task. Since some pre-training objectives naturally produce long outputs by design, we enforce generation length constraints in the zero-shot setting to ensure a reasonable comparison of model outputs, as done in \citet{primera}. Table \ref{tab:model_summaries} displays example zero-shot outputs from each model.\\

\MethodName{} demonstrates strong zero-shot performance, consistently outperforming the baselines on most metrics. First, \MethodName{} outperforms \PrimeraName{} and Pegasus-X on Rouge-G by an average of 1.7 and 3.1 points, and on BertScore by 1.2 and 2.1 points, respectively. We further observe improved coherence (DiscoScore) and faithfulness (MDSummaC), all while achieving significantly higher summary abstractiveness for all domains. 

Of the baselines, QAmden struggles the most, especially in the zero-shot setting, largely due to its QA-centric pre-training objective which may imbue strong cross-document capabilities, but is not well-suited to summarization without fine-tuning. We observe the outputs are largely incoherent and somewhat in the format of answers to questions, which aligns with its pre-training.

Centrum is the strongest baseline, achieving solid performance, particularly on the news domains; we see improved Rouge and BertScore performance over PELMS from Centrum on Multi-News and DUC2004, however, results are mixed for Multi-XScience and noticeably worse on the opinion summarization datasets, with PELMS performing much better. We see coherence (DiscoScore) is comparable between Centrum and PELMS, indicating that our coherence heuristics achieve performance comparable with a technique whose leave-one-out whole-document target is naturally coherent.

Notably, PELMS achieves strong faithfulness (MDSummac) over all domains, speaking to the strength of our pre-training objective which enforces input-consistency selection constraints at the sentence granularity. In contrast, faithfulness is somewhat worse for Centrum, perhaps as its objective selects an entire document as the proxy summary, and it therefore becomes heavily reliant on whether the task contains standalone documents that are both informative and faithful to the rest of the input. Our sentence-granularity selection method is more flexible and can better adapt to tasks where these assumptions do not hold.

We note that Centrum was pre-trained (using the NewSHead dataset) and primarily evaluated on \textit{news} data, a domain very suitable for their leave-one-out pre-training objective due to the large content overlap across news articles. \citet{wolhandler2022multi_how_multi_is_mds} note that news datasets often require little to no cross-document synthesis for strong performance, thus our emphasis on expanding evaluation to other domains, where a single document cannot realistically represent the input.

\begin{table*}[t]
    \centering
    \small
    \include{tables/model_ablation}
    \caption{
   We pre-train with the \PrimeraName{} technique using our MultiPT data, and pre-train our \MethodName{} technique with the \PegasusXName{} long-input architecture (initializing our model with \PegasusXName{} weights). We report the overall zero-shots results.}
    \label{tab:modelablation}
\end{table*}

\begin{table}[t]
    \centering
    \small
    \setlength{\tabcolsep}{2pt}
    \include{tables/human_eval}
    \caption{Human evaluation on MetaTomatoes dataset. For Grammaticality, Referential Clarity, and Structure \& Coherence, we report the average ranking when comparing methods. Ties are allowed. For Faithfulness, we report the percentage of system-generated SCUs that are holistically entailed by human-verified input SCUs. We achieve statistically significant values in Referential Clarity, Structure \& Coherence, and Faithfulness (p < 0.05).}
    \label{tab:human_eval}
\end{table}

\subsection{Supervised Evaluation}
In the supervised analysis, we explore both full parameter tuning and parameter-sparse updates using adapters (training only 5\% of model parameters). We evaluate on 16-shot, 64-shot, and full-shot data quantities. All experiments are averaged over 5 runs with unique seeds. Table \ref{tab:full_main_results} shows overall results averaged over the six datasets. Per-dataset results w/ statistical significance tests are provided in Appendix \ref{appx:full_results_stat_sig}.

\paragraph{Full-parameter tuning}

As expected, we see summary quality improves with further supervision. \MethodName{} obtains the highest ROUGE-G improvements across the board, notably improving ROUGE-G by 1.2 points on the 16-shot split. For larger quantities, we see some convergence on both ROUGE and BertScore, with \MethodName{} outperforming \PrimeraName{} by 0.5 points on ROUGE-G and 0.6 points on BertScore for both 64-shot and full-shot splits. With respect to summary characteristics, \PegasusXName{} achieves very high abstractiveness (N-gram Novelty) with full supervision, yet the lowest faithfulness. 

Even with full-parameter updates, QAmden suffers in few-shot setups, with noticeably lower ROUGE, BertScore and DiscoScore. With full shot and full supervision, however, it nears the performance of the other methods. 

Centrum fares worse than others on full supervision, with just 2.5 point BertScore improvement from zero-shot to full-shot, and fairly minimal improvements with few-shot training. In contrast, \MethodName{} achieves a 4.1 point improvement with full supervision. \textit{Full-parameter tuning with \MethodName{} achieves the highest ROUGE and BertScore at all training data splits}. 

\paragraph{Adapter fine-tuning}
We see \MethodName{} \textit{significantly outperforms all other methods when performing adapter fine-tuning}. In 16-shot, we see 2.4 points of ROUGE-G improvement over \PrimeraName{}, and 5.4 over \PegasusXName{}, with similar trends in 64-shot and full-shot splits with Centrum and QAmden as well. For all of the models, adapter training tends to converge to lower ROUGE and BertScore values, while maintaining a more moderate balance between abstractiveness and faithfulness. Discourse scores are fairly consistent accross most models, and generally correlated with informativeness. 

Overall, we see that \MethodName{} is strong in both fully-supervised and PEFT settings, achieving the highest summary quality, while maintaining strong informativeness across the secondary summary characteristics. In particular, the PEFT results highlight our method's ability to adapt effectively to data and compute-scarce environments, validating the alignment between our pre-training objective and the task at hand.

\section{Further Analyses}
\label{sec:analyses}

\paragraph{Comparison with LLMs}
We briefly compare performance between the pre-trained MDS models and state-of-the-art LLMs. We report the overall results using both GPT-3.5-Long (16k tokens) and GPT-4 (8k tokens) in a zero-shot setting. We use the 0613 versions for both. For a fair comparison, inputs are all truncated to the same 4,096 tokens as supported by the baseline models. We follow the MDS prompts used by \citet{caciularu-etal-2023-peek-qamden}, with a simple instruction to generate a multi-document summary for following input. We report the overall results in Table \ref{tab:full_main_results} and the full results in Appendix 
\ref{appx:llm_results}. We see strong performance, particularly in informativeness, coherence and abstractiveness. We note that \MethodName{} \textit{achieves comparable or better overall performance in as few as 16 shots versus GPT-3.5-Long and in 64 shots versus GPT-4.0 for the ROUGE, BertScore and DiscoScore metrics.} Additionally we observe abstractiveness (N-gram Novelty) is very high for the GPT models compared to the baselines, although this is contrasted with relatively low MDSummaC faithfulness.

\paragraph{Pre-training Ablation.}
We investigate the individual benefits of both our \PretrainDatasetName{} data and \MethodName{} pre-training objective on other models (Table \ref{tab:modelablation}). We find \PrimeraName{} benefits from pre-training on our diverse large-scale dataset (as opposed to NewsHead). Additionally, we see strong performance when initializing our method with \PegasusXName{} instead of LED; this variant outperforms the LED model on most metrics, although faithfulness is decreased. These differences in performance may be due to the architecture differences, or perhaps due to the differences in the existing pre-training the base models had already undergone.

\paragraph{Human Evaluation.}
We additionally evaluate the quality of our method with human judges, using three human judges for fluency evaluation and two for faithfulness evaluation. We follow \citet{primera} in evaluating both summary fluency and faithfulness, performing our evaluation on 25 examples from the MetaTomatoes dataset due to the time-consuming nature of the in-depth scoring process, especially for the faithfulness evaluation which requires extensive comparison of input and output. Our model showed improved grammaticality, referential clarity, structure, and coherence compared to the baseline GSG-style pre-training methods. We also find our model has highest input faithfulness, generating summaries that were the most reflective of their inputs. Appendix \ref{appx:human_eval} contains further human evaluation details.

\paragraph{Length Control Experiment.}
We briefly explore length control during pre-training, varying the $k$ value which sets the number of target sentences and training with a corresponding a length-prefix. We achieve an average ROUGE-G improvement of 0.6 points. Further details and results of this experiment can be found in Appendix \ref{appx:length_control}.
\\\\\\

\section{Conclusion}
We introduce \MethodName{}, a novel new multi-document summarization pre-training method. \MethodName{} leverages semantic topic clustering to improve summary informativeness and uses coherence and faithfulness constraints during pre-training target formulation to produce concise, fluent, and faithful summaries. Both automatic evaluation and human
evaluation demonstrate that \MethodName{} achieves consistent improvements over competitive baseline pre-trained models, yielding especially strong performance in low-shot and parameter-efficient training settings.

\section*{Limitations}
As our emphasis was on building a proficient pre-trained model, our technique is complementary to fine-tuning-specific methods that attempt to further inject specific summarization constraints during fine-tuning. For example, inference-time or train-time techniques such as FactPEGASUS \cite{wan-bansal-2022-factpegasus} and CLIFF \cite{cao2021cliff} can still be used during supervised fine-tuning with our base model.

\bibliography{acl_latex}
\bibliographystyle{acl_natbib}

\appendix

\section{\MethodName{} Pre-training Details}
\label{appx:pretraining_technique_details}

\subsection{\MethodName{} Technique Configuration}

\paragraph{Topic Clustering}
For semantic topic clustering, we use the Sentence Transformers \textit{all-mpnet-base-v2} embedding model, which is derived from \citet{song2020mpnet}. We cluster using the Sentence Transformers community-based `fast clustering' algorithm, using cosine similarity as the distance metric. We specify a distance threshold of 0.6, and only consider clusters with at least 2 elements. \\
As described in \cref{sec:technique}, \MethodName{} topic selection is parameterized by \textit{k}, enabling us to vary the length per example. In half of the training examples, we set k to 8 for enabling general purpose summarization. For the remaining examples, we sample k to enable length-controlled summarization, as detailed in Appx. \ref{appx:length_control}. \\

\paragraph{Sentence Ranking + Selection}
For the NLI entailment step, we use the \textit{albert-base-vitaminc} \cite{schuster-etal-2021-get-vitac} model. We use only the positive entailment probability when calculating intra-cluster entailment.
Additionally, the ranking and ordering steps are parametrized by $c$, which is the number of elements considered for selection from each topic cluster. We set this to 2 in our pre-training.

\subsection{Additional Pre-training Details}
\label{appx:additional_pretraining_details}

We train \MethodName{} from the 464M LED-large base model, using the default sliding-window local attention configuration supplemented by inserting \texttt{<doc-sep>} tokens with full attention added after each document. For all examples, we cap the input sizes at 4,096 tokens to enable tractable training. This cap affects approximately 10\% of pre-training inputs. When truncating long inputs, we distribute the tokens equally among the documents. \\
We train on 16 GPUs with a total effective batch size of 1,024, learning rate of 5e-5, and 1,250 warmup steps. Leveraging 16 NVIDIA A40 GPUs, this takes approximately 4 days. We train over the full pre-training dataset for one epoch. Pre-processing of the 3-million+ pre-training examples with the \MethodName{} technique takes approximately 14 hours using the same computing environment. \\
We perform basic dataset pre-processing of \PretrainDatasetName{} prior to pre-training target creation, removing extremely short and/or noisy documents containing HTML content.

\section{Evaluation Details}
\label{sec:experiment_details_appendix}

\paragraph{Input Preparation}
For fair comparison of all models, we pre-process the input documents similarly, with input sizes capped at 4,096 tokens during training and evaluation. If truncation is necessary, we distribute the tokens equally among the documents, truncating the end of each document.

\paragraph{Generation Settings}
Largely following the existing methods, for all experiments, we use beam search as the generation decoding strategy, with the number of beams set to 5. We enable tri-gram blocking during decoding. Following \citet{primera}, we set a maximum generation length per dataset to ensure reasonable and comparable output lengths. For Amazon and Yelp we use 96, DUC2004 and Multi-XScience we use 128, MetaTomatoes we use 192, and MultiNews we use 256.

\paragraph{Model Training}
We use a learning rate of 3e-4 in our experiments, training for a maximum of 30 epochs. We base this off of hyperparameters from the the four baseline models which all leverage the same LED-large model. We apply early stopping based on validation Rouge-G score. For few-shot experiments, we scale down the validation set size to match the few-shot size (for example, in 16-shot training, we use a validation set of 16 examples). Results are averaged over 5 runs. For datasets with multiple ground-truth references, we unpack them and treat each reference as a separate example or ``shot''.

\paragraph{Test Sets}
We cap test set size to 1,000 examples to ensure tractable evaluation as the neural DiscoScore and MDSummaC automated metrics used in our evaluation are GPU compute-intensive. This impacts only the  Multi-XScience and Multi-News evaluation. MDSummaC takes approximately 45 minutes to evaluate 1,000 examples. 

\paragraph{Model Selection}
Due to computational limitations, we evaluate only on recent strong baselines: \PrimeraName{} and \PegasusXName{}, Centrum, and QAMDen. These outperform other baselines like LED and BART \cite{lewis2019bart}. We explored Flan-T5, a T5-based method pre-trained for multiple tasks \cite{flan-t5}, but it showed poor generalization to multi-document summarization in our pilot studies, likely due to lack of long-input pre-training.

\section{Dataset Information}

\subsection{Pre-training Datasets}
The datasets used in our \PretrainDatasetName{} corpus are described here. 
\begin{itemize}
    \item NewSHead \cite{gu2020generating_newshead} - A dataset of news stories published between between 2018-2019, grouped together to form topic-centric document clusters.
    \item BigNews-Aligned \cite{liu-etal-2022-politics-bignews} - A large-scale news dataset containing over 3 million english political news articles which are clustered by event. These articles are gathered from 11 large US news sites.
    \item WikiSum-40 \cite{wikisum40} - Derived from WikiSum \cite{liu2018generating_wikisumog} a Wikipedia corpus containing wikipedia articles and their source articles. WikiSum-40 is a filtered variant containing only the 40 most-relevant documents for every wikipedia article.
    \item AmazonPT \cite{ni-etal-2019-amazon} - A subset of the massive dataset of Amazon product reviews. We use a 1,000,000 cluster subset of this, sampling uniformly from the available categories.
    \item YelpPT (https://www.yelp.com/dataset) - a reviews dataset containing consumer reviews of businesses.
\end{itemize}

\subsection{Evaluation Datasets}
We provide further details on the datasets used during evaluation, and introduce our new \RottenTomatoesName{} dataset.
\label{appx: datasets_extra}

\begin{table*}[t]
\centering
    \small
    \include{tables/eval_datasets}
    \caption{Overview of the six datasets used in our MDS evaluation.
    }
    \label{tab:eval_datasets}
    \vspace{-3mm}
\end{table*}

\begin{itemize}
    \item Multi-News \cite{fabbri2019multinews} - A large-scale MDS news summarization dataset containing 56,216 articles-summary pairs.
    \item DUC2004 \cite{duc2004_dang2005overview} - A small carefully-curated news summarization dataset containing 50, 10-document news article clusters and corresponding summaries.
    \item Multi-XScience \cite{lu2020multixscience} - A related-work generation task, with the goal of consolidating information from scientific article abstracts cited by a given paper.
    \item Amazon, Yelp \cite{amazon_yelp_bravzinskas2020few} - Two small crowdworker-curated opinion summarization datasets with focus on aggregating opinions expressed in consumer reviews of consumer products (Amazon), and businesses (Yelp).
\end{itemize}

\subsubsection{MetaTomatoes Meta-review Dataset}
\begin{figure}[t]
\centering
     \includegraphics[width=\columnwidth]{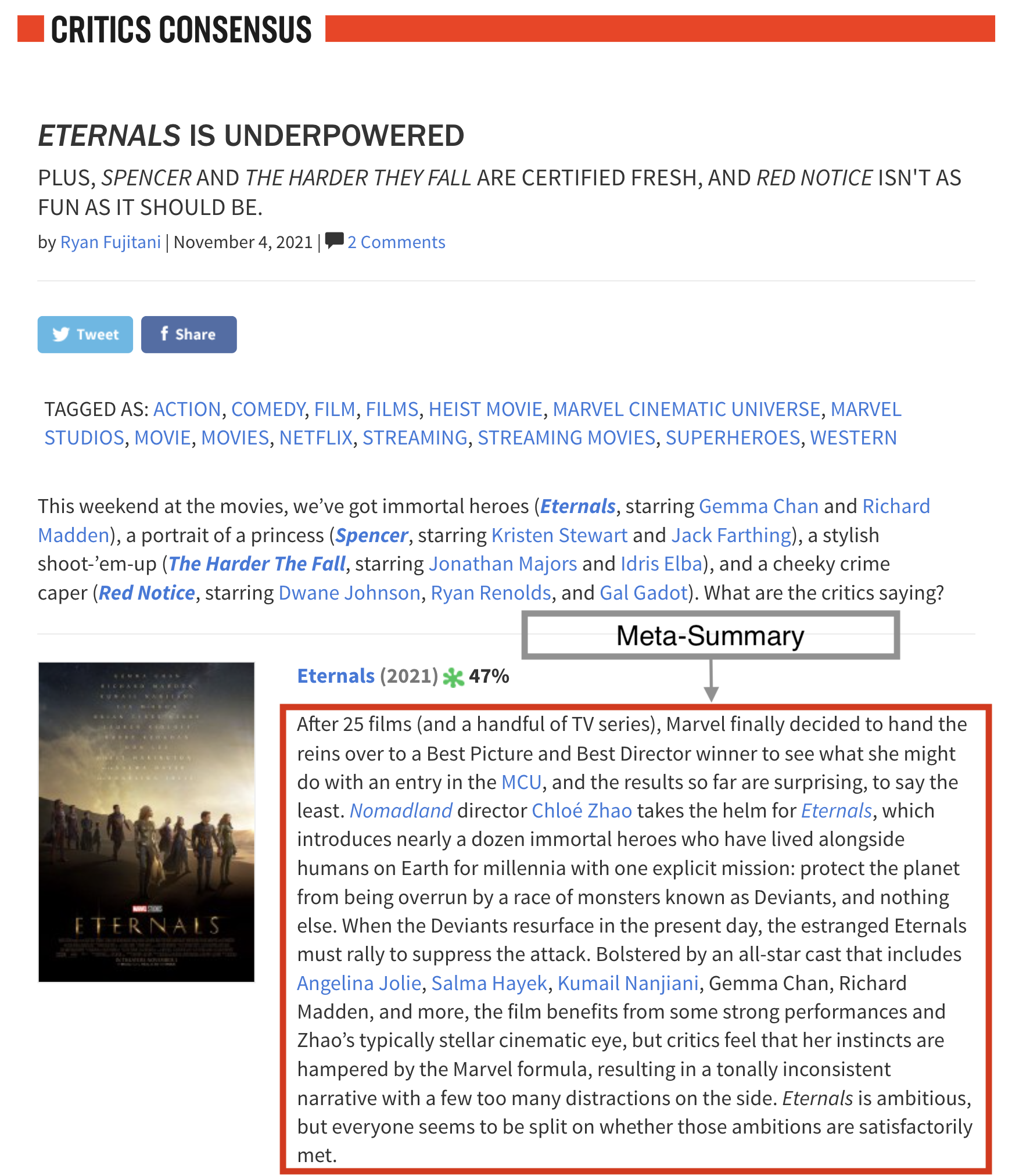}
      \caption{Example Rotten Tomatoes Critics Consensus Meta-Summary}
     \label{fig:meta_tomatoes_source}
\end{figure}

\citet{rt_v1_wang-ling:2016:N16-1} previously released RottenTomatoes, an MDS meta-review dataset in which inputs consisted of many short editorial summaries produced by RottenTomatoes contributors. The objective was to generate a brief sentence that captures the overall opinion towards the new film.
In contrast, we identify and scrape longer-form meta-reviews produced by the RottenTomatoes editorial team using similar inputs. Figure \ref{fig:meta_tomatoes_source} displays an example meta-summary. These meta-reviews are significantly longer than any one input ‘contributor review summary’, with the input posing a unique challenge due to the large number of documents, necessitating cross-document information aggregation and understanding.

\section{Length-Controlled Summarization}
As mentioned in Section \ref{sec:analyses}, we perform length-controlled summarization experimentation in the zero-shot setting. During pre-training we train with a fixed $k$ of 8 for half of the examples. For the remainder, we randomly sample k from a normal distribution (mean=7, std\_dev=5) to encourage flexibility within the model output; in these cases, we prepend the input with a corresponding length prefix, corresponding to one of five length bins. We bound k within the range [1,14]. Table \ref{tab:controlled_results} overviews the results and best prefixes for each dataset.

The five length prefixes and corresponding bins are as follows:
\begin{itemize}
    \item `short': [1,2]
    \item `short-medium': [3,5]
    \item `medium': [6,8]
    \item `medium-long': [9,11]
    \item `long': [12, 14]
\end{itemize}

\label{appx:length_control}
\begin{table*}[t]
    \centering
    \small
    \include{tables/control_results}
    \caption{Comparison of uncontrolled vs length-controlled zero-shot performance. The controlled setting varies only in the prefix supplied during inference. We report the best prefix per dataset, noting that the best prefixes aligned well with the expected best (i.e., longer prefixes for datasets with longer summaries). Our selection metric is RG score.}
    \label{tab:controlled_results}
\end{table*}

\section{Additional Evaluation Metrics Details}
\label{appx:eval_metrics}

We provide additional information on the 5 metrics used in our evaluation. All metrics report values in range [0,1] with displayed results multiplied by 100 for presentation purposes. For all, higher scores are better.

\paragraph{Summary Informativeness}
\begin{itemize}
    \item ROUGE \cite{lin2004rouge}, an n-gram overlap metric commonly used for summarization evaluation.
    \item BertScore \cite{zhang2019bertscore}, a BERT-based text similarity metric, also popular for text generation evaluation.
\end{itemize}

\paragraph{Coherence}
\begin{itemize}
    \item DiscoScore \cite{discoscore}, a recent BERT-based method for evaluating discourse coherence. In particular, we use the DS-SENT (NN) variant---which is shown to correlate well with human judgment, particularly in the news domain---to measure the discourse similarities between system and reference summaries.
\end{itemize}
\paragraph{Faithfulness}
\begin{itemize}
    \item MDSummaC - To better measure consistency between multiple-document inputs and a system summary,\textbf{ we introduce MDSummaC}, an entailment-based consistency metric that we extend from the SummaC \cite{laban2021summac} consistency metric. SummaC uses entailment to identify whether a text is consistent with another. In particular, we repurpose the $SummaC_{ZS}$ variant, which uses sentence-wise comparison of the input text and generated summary, calculating consistency as the maximum entailment score between a given summary sentence and any input sentence, then reporting the average of this over all of the summary sentences. Unfortunately, this formulation struggles to fit the multi-document case, as SummaC will return a high score even if a summary is maximally consistent with sentences from just one document.
    To ensure the summary is independently consistent with each document, we instead report the average $SummaC_{ZS}$ score over each individual input document $Doc_i$ in the input $I$ as follows:
    \begin{equation}
    \small
    \text { MDSummaC }(I,S)=\frac{1}{n} \sum_{i=1}^{n} \operatorname{SummaC_{ZS}}\left(Doc_i, S\right)
    \end{equation}
\end{itemize}
\paragraph{Abstractiveness}
\begin{itemize}
    \item N-gram novelty - Calculates the abstractiveness of a summary as the proportion of novel n-grams within the summary. For example, unigram novelty captures the proportion of summary unigrams not seen in the input. To simplify our reporting, we report the arithmetic mean of 1-gram, 2-gram and 3-gram novelty as our proxy for summary abstractiveness.
\end{itemize}

\section{Human Evaluation Details}
We hire 3 native English speakers to act as human evaluators of generated summaries. We use \cite{primera}'s guidelines for fluency evaluation, measuring grammaticality, referential clarity, and structure \& coherence. As their instructions confusingly mixed both absolute scoring with suggestions to perform comparative scoring, we simplified the task to a comparative ranking of the system outputs from our three evaluated models, with ties allowed. 

Summaries were presented to annotators in random order. We performed the faithfulness evaluation in two phases. First, in the filtering phase, annotators were asked to identify all Summary Content Units (SCUs) within the inputs. We used majority vote to merge these annotations. Next, once SCUs had been selected, the annotators were provided the system summaries and asked to score each each summary sentence as either 1 or 0, with 1 meaning the sentence was holistically entailed by the input SCUs. To produce an overall score, we report the average percentage of summary sentences that were entailed by the input SCUs. 

\begin{figure*}[htbp!]
    \centering
    \includegraphics[width=\textwidth]{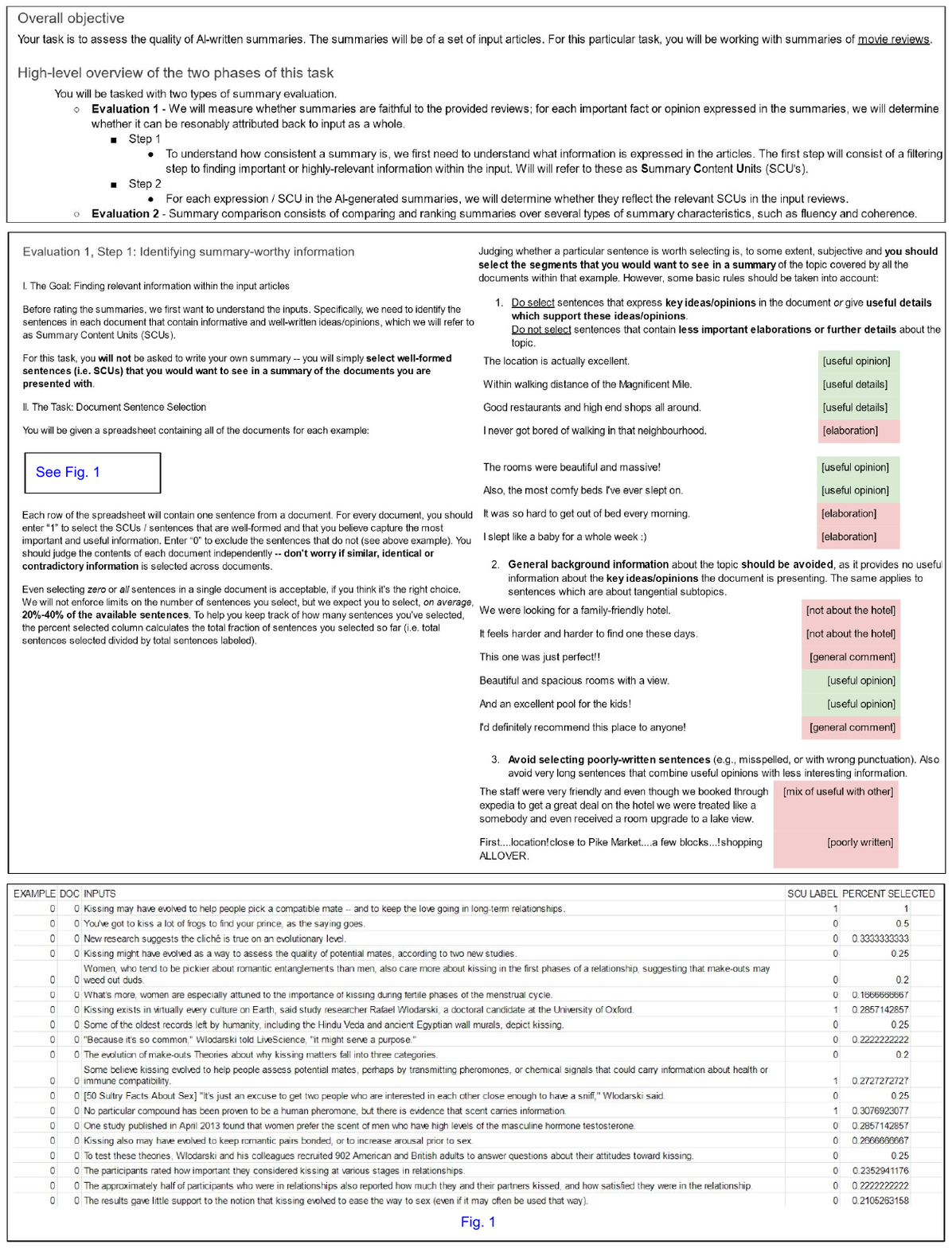}
    \caption{Human Evaluation Guidelines Pt. 1}
    \label{fig:human_eval_1}
  \end{figure*}

\begin{figure*}[htbp!]
    \centering
    \includegraphics[width=\textwidth]{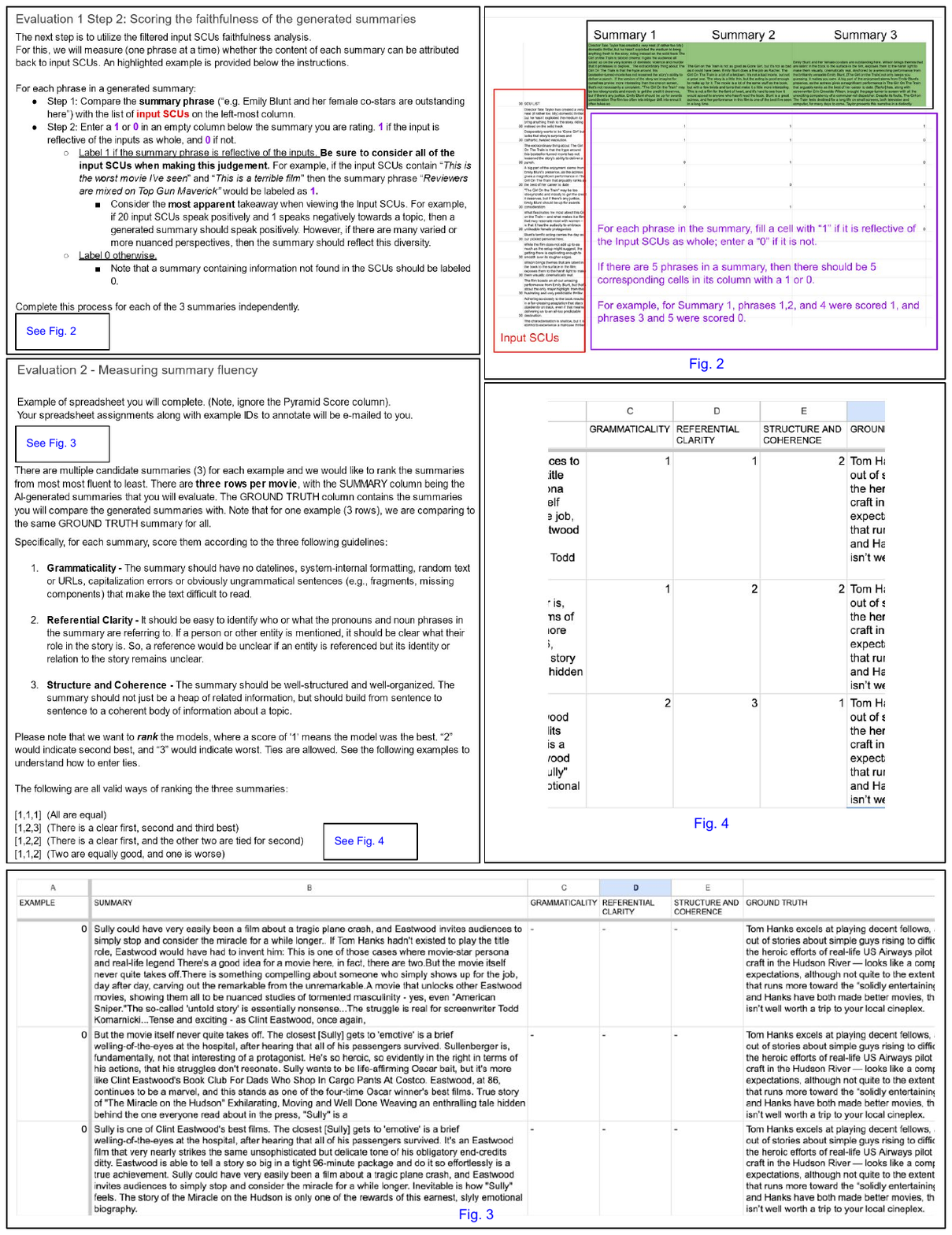}
    \caption{Human Evaluation Guidelines Pt. 2}
    \label{fig:human_eval_2}
  \end{figure*}

Figures \ref{fig:human_eval_1} and \ref{fig:human_eval_2} contain the guidelines provided during the human evaluation. Our inter-annotator agreement scores were 0.43 for Grammaticality, 0.53 for Referential Clarity, 0.28 for Structure \& Coherence, and 0.47 for Faithfulness.
\label{appx:human_eval}

\section{Full LLM Results}
\label{appx:llm_results}
Table \ref{tab:llm_results} outlines the results of GPT-3.5-Long and GPT-4 on each of the six evaluation datasets.

\section{Full Supervised Results}
\label{appx:full_results_stat_sig}
We provide the full results for all datasets with our supervised setups (all combinations of 16/64/full-shot splits and full-parameter/adapter training) in Tables \ref{tab:16_shot_full_param}, \ref{tab:16_shot_adapter}, \ref{tab:64_shot_full_param}, \ref{tab:64_shot_adapter}, \ref{tab:full_shot_full_param}, and \ref{tab:full_shot_full_adapter}.

\begin{table*}
    \centering
    \tiny
    \include{tables/model_summaries}
    \caption{Zero-shot MultiNews Summary Output Example}
    \label{tab:model_summaries}
\end{table*}

\begin{table*}[t]
    \centering
    \small
    \include{tables/_llm_full}
    \caption{Results of LLMs on the multi-document summarization task. We see GPT-4 achieves consistently higher results compared to GPT-3.5-Long. In particular, GPT-4 is 1.2 points better than GPT-3.5 for BertScore informativeness. Coherence, faithfulness, and abstractiveness are also moderately improved. }
    \label{tab:llm_results}
\end{table*}

\clearpage

  \clearpage

\clearpage

\begin{table*}[t]
    \centering
    \tiny
    \include{tables/16_full_parameter}
    \caption{16-shot results with full-parameter training. 	In our experiments, we average over 5 runs, each with unique random seeds. We report the mean and (std) values. \colorbox[HTML]{B7E1CD}{Green} and \^{} indicate PELMS outperforms all baselines. \\\textbf{Bold} and \text{*} indicate improvement is statistically significant (one-tailed paired t-test with each baseline, p < 0.05).}
    \label{tab:16_shot_full_param}
\end{table*}

\begin{table*}[t]
    \centering
    \tiny
    \include{tables/16_adapter}
    \caption{16-shot results with adapter (5\%) training. In our experiments, we average over 5 runs, each with unique random seeds. We report the mean and (std) values. \colorbox[HTML]{B7E1CD}{Green} and \^{} indicate PELMS outperforms all baselines. \\\textbf{Bold} and \text{*} indicate improvement is statistically significant (one-tailed paired t-test with each baseline, p < 0.05).}
    \label{tab:16_shot_adapter}
\end{table*}

\begin{table*}[t]
    \centering
    \tiny
    \include{tables/64_full_parameter}
    \caption{64-shot results with full-parameter training. In our experiments, we average over 5 runs, each with unique random seeds. We report the mean and (std) values. \colorbox[HTML]{B7E1CD}{Green} and \^{} indicate PELMS outperforms all baselines. \\\textbf{Bold} and \text{*} indicate improvement is statistically significant (one-tailed paired t-test with each baseline, p < 0.05).}
    \label{tab:64_shot_full_param}
\end{table*}

\begin{table*}[t]
    \centering
    \tiny
    \include{tables/64_adapter}
    \caption{64-shot results with adapter (5\%) training. In our experiments, we average over 5 runs, each with unique random seeds. We report the mean and (std) values. \colorbox[HTML]{B7E1CD}{Green} and \^{} indicate PELMS outperforms all baselines. \\\textbf{Bold} and \text{*} indicate improvement is statistically significant (one-tailed paired t-test with each baseline, p < 0.05).}
    \label{tab:64_shot_adapter}
\end{table*}

\begin{table*}[t]
    \centering
    \tiny
    \include{tables/full_shot_full_parameter}
    \caption{Full-shot results with full-parameter training. In our experiments, we average over 5 runs, each with unique random seeds. We report the mean and (std) values. \colorbox[HTML]{B7E1CD}{Green} and \^{} indicate PELMS outperforms all baselines. \\\textbf{Bold} and \text{*} indicate improvement is statistically significant (one-tailed paired t-test with each baseline, p < 0.05).}
    \label{tab:full_shot_full_param}
\end{table*}

\begin{table*}[t]
    \centering
    \tiny
    \include{tables/full_shot_adapter}
    \caption{Full-shot results with adapter (5\%) training. In our experiments, we average over 5 runs, each with unique random seeds. We report the mean and (std) values. \colorbox[HTML]{B7E1CD}{Green} and \^{} indicate PELMS outperforms all baselines. \\\textbf{Bold} and \text{*} indicate improvement is statistically significant (one-tailed paired t-test with each baseline, p < 0.05).}
    \label{tab:full_shot_full_adapter}
\end{table*}

\end{document}

%% file: tables/pretraining_data.tex
\begin{tabular}{lcrrrrr}
\hline
\textbf{MultiPT Datasets} & \textbf{Domain}      & \multicolumn{1}{c}{\textbf{\#Clusters}} & \multicolumn{1}{c}{\textbf{\#Docs / C}} & \multicolumn{1}{c}{\textbf{Doc\_len}} & \multicolumn{1}{c}{\textbf{Input\_len}} & \multicolumn{1}{c}{\textbf{Total \#Tokens}} \\ \hline
Newshead                              & News                 & 400,000                                 & 3.5                                     & 495.2                                 & 1732                                    & 1,733M                                      \\
BigNews-Aligned                       & News                 & 1,060,512                               & 4.3                                     & 656.1                                 & 2820                                    & 2,821M                                      \\
WikiSum-40                            & Wikipedia            & 1,000,000                               & 40                                      & 70.1                                  & 2804                                    & 2,800M                                      \\
AmazonPT                              & Product Reviews      & 1,000,000                               & 40.3                                    & 72.8                                  & 2901                                    & 2,970M                                      \\
YelpPT                                & Business Reviews     & 142,000                                 & 61.3                                    & 60.6                                  & 3714                                    & 528M                                        \\ \hline
\end{tabular}

%% file: tables/_full_zero_shot_analysis.tex
\begin{tabular}{ll|cccccccc}
\hline
\textbf{Dataset} & \textbf{Model} & \textbf{R1}                           & \textbf{R2}                          & \textbf{RL}                           & \textbf{RG}                           & \textbf{BertS}                        & \textbf{DiscoS}                       & \textbf{MDSummaC}                     & \textbf{N-gram Novelty} \\ \hline
MultiNews        & Pegasus-X      & 39.1                                  & 11.2                                 & 17.8                                  & 19.8                                  & 59.2                                  & 91.5                                  & 37.8                                  & 4.6                     \\
                 & Primera        & 41.9                                  & 12.7                                 & 19.5                                  & 21.8                                  & 60.8                                  & 93.1                                  & \textbf{41.3}                         & 3.6                     \\
                 & QAmden         & 31.1                                  & 7.9                                  & 15.7                                  & 15.6                                  & 49.9                                  & 63.5                                  & 5.3                                   & \textbf{40.2}           \\
                 & Centrum        & \textbf{44.2}                         & \textbf{15.1}                        & \textbf{21.6}                         & \textbf{24.3}                         & \textbf{62.2}                         & \textbf{95.7}                         & 39.2                                  & 11.5                    \\
                 & PELMS          & 41.7                                  & 13.5                                 & 20.0                                  & 22.4                                  & 61.0                                  & 95.3                                  & 38.6                                  & 7.8                     \\ \hline
Multi-XScience   & Pegasus-X      & 28.8                                  & 4.5                                  & 14.9                                  & 12.5                                  & 55.7                                  & 89.8                                  & 30.5                                  & 3.3                     \\
                 & Primera        & 29.6                                  & 4.6                                  & 15.2                                  & 12.8                                  & 55.5                                  & 81.7                                  & 27.0                                  & 1.6                     \\
                 & QAmden         & 25.4                                  & 3.5                                  & 14.3                                  & 10.9                                  & 51.9                                  & 71.9                                  & 9.6                                   & \textbf{22.8}           \\
                 & Centrum        & 29.2                                  & 4.6                                  & 15.3                                  & 12.7                                  & 55.2                                  & 90.8                                  & \textbf{31.5}                         & 7.2                     \\
                 & PELMS          & \cellcolor[HTML]{B1CBF2}\textbf{30.0} & \cellcolor[HTML]{B1CBF2}\textbf{4.8} & \cellcolor[HTML]{B1CBF2}\textbf{15.4} & \cellcolor[HTML]{B1CBF2}\textbf{13.0} & \cellcolor[HTML]{B1CBF2}\textbf{56.0} & \cellcolor[HTML]{B1CBF2}\textbf{90.9} & 31.1                                  & 3.0                     \\ \hline
Amazon           & Pegasus-X      & 27.2                                  & 4.2                                  & 15.1                                  & 12.0                                  & 58.0                                  & \textbf{90.8}                         & \textbf{15.3}                         & 5.1                     \\
                 & Primera        & 28.1                                  & 5.1                                  & 16.4                                  & 13.3                                  & 59.0                                  & 84.1                                  & 8.1                                   & 2.5                     \\
                 & QAmden         & 24.1                                  & 3.3                                  & 14.5                                  & 10.5                                  & 54.9                                  & 77.2                                  & 6.2                                   & 16.2                    \\
                 & Centrum        & 28.9                                  & 4.8                                  & 16.8                                  & 13.3                                  & 58.5                                  & 90.4                                  & 12.3                                  & 25.3                    \\
                 & PELMS          & \cellcolor[HTML]{B1CBF2}\textbf{31.3} & \cellcolor[HTML]{B1CBF2}\textbf{7.1} & \cellcolor[HTML]{B1CBF2}\textbf{18.7} & \cellcolor[HTML]{B1CBF2}\textbf{16.1} & \cellcolor[HTML]{B1CBF2}\textbf{62.0} & \cellcolor[HTML]{B1CBF2}\textbf{90.6} & 12.8                                  & \textbf{30.8}           \\ \hline
Yelp             & Pegasus-X      & 25.0                                  & 3.9                                  & 14.3                                  & 11.2                                  & 59.0                                  & 89.9                                  & \textbf{14.5}                         & 6.9                     \\
                 & Primera        & 27.4                                  & 4.9                                  & 15.7                                  & 12.8                                  & 59.8                                  & 85.2                                  & 10.7                                  & 2.7                     \\
                 & QAmden         & 21.2                                  & 3.3                                  & 13.2                                  & 9.7                                   & 54.8                                  & 75.7                                  & 6.9                                   & 17.5                    \\
                 & Centrum        & 27.4                                  & 5.4                                  & 16.7                                  & 13.5                                  & 58.8                                  & 89.9                                  & 9.8                                   & \textbf{39.1}           \\
                 & PELMS          & \cellcolor[HTML]{B1CBF2}\textbf{31.7} & \cellcolor[HTML]{B1CBF2}\textbf{8.0} & \cellcolor[HTML]{B1CBF2}\textbf{19.0} & \cellcolor[HTML]{B1CBF2}\textbf{16.9} & \cellcolor[HTML]{B1CBF2}\textbf{62.4} & \cellcolor[HTML]{B1CBF2}\textbf{91.6} & 14.2                                  & 24.9                    \\ \hline
DUC2004          & Pegasus-X      & 31.5                                  & 5.0                                  & 15.5                                  & 13.5                                  & 56.4                                  & 92.8                                  & 4.6                                   & 4.1                     \\
                 & Primera        & 32.9                                  & 7.0                                  & 16.9                                  & 15.8                                  & 58.1                                  & 77.1                                  & 12.6                                  & 2.3                     \\
                 & QAmden         & 28.2                                  & 4.3                                  & 15.3                                  & 12.3                                  & 53.8                                  & 75.7                                  & 8.9                                   & \textbf{12.1}           \\
                 & Centrum        & \textbf{34.7}                         & \textbf{7.9}                         & \textbf{18.2}                         & \textbf{17.1}                         & \textbf{59.3}                         & \textbf{94.0}                         & 13.8                                  & 3.4                     \\
                 & PELMS          & 34.0                                  & 6.6                                  & 16.7                                  & 15.6                                  & 58.8                                  & 93.0                                  & \cellcolor[HTML]{B1CBF2}\textbf{15.4} & 7.0                     \\ \hline
MetaTomatoes     & Pegasus-X      & 31.1                                  & 4.6                                  & 13.8                                  & 12.5                                  & 54.4                                  & 93.4                                  & 5.4                                   & 12.5                    \\
                 & Primera        & 32.1                                  & 5.0                                  & 14.4                                  & 13.2                                  & 54.9                                  & 80.5                                  & 3.5                                   & 7.1                     \\
                 & QAmden         & 21.6                                  & 2.6                                  & 11.6                                  & 8.6                                   & 44.5                                  & 72.3                                  & 1.9                                   & \textbf{70.2}           \\
                 & Centrum        & 32.5                                  & 5.6                                  & 15.0                                  & 13.9                                  & 54.9                                  & \textbf{93.4}                         & 4.7                                   & 26.9                    \\
                 & PELMS          & \cellcolor[HTML]{B1CBF2}\textbf{34.1} & \cellcolor[HTML]{B1CBF2}\textbf{7.4} & \cellcolor[HTML]{B1CBF2}\textbf{16.5} & \cellcolor[HTML]{B1CBF2}\textbf{16.1} & \cellcolor[HTML]{B1CBF2}\textbf{55.1} & 92.0                                  & \cellcolor[HTML]{B1CBF2}\textbf{6.2}  & 41.4                    \\ \hline
Overall          & Pegasus-X      & 30.4                                  & 5.6                                  & 15.2                                  & 13.6                                  & 57.1                                  & 91.4                                  & 18.0                                  & 6.1                     \\
                 & Primera        & 32.0                                  & 6.6                                  & 16.3                                  & 15.0                                  & 58.0                                  & 83.6                                  & 17.2                                  & 3.3                     \\
                 & QAmden         & 25.3                                  & 4.1                                  & 14.1                                  & 11.3                                  & 51.6                                  & 72.7                                  & 6.5                                   & \textbf{29.8}           \\
                 & Centrum        & 32.8                                  & 7.2                                  & 17.3                                  & 15.8                                  & 58.2                                  & \textbf{92.4}                         & 18.6                                  & 18.9                    \\
                 & PELMS          & \cellcolor[HTML]{B7E1CD}\textbf{33.8} & \cellcolor[HTML]{B7E1CD}\textbf{7.9} & \cellcolor[HTML]{B7E1CD}\textbf{17.7} & \cellcolor[HTML]{B7E1CD}\textbf{16.7} & \cellcolor[HTML]{B7E1CD}\textbf{59.2} & 92.2                                  & \cellcolor[HTML]{B7E1CD}\textbf{19.7} & 19.2                    \\ \hline
\end{tabular}

%% file: tables/_overall_summary.tex
\begin{tabular}{l|lcccccccc}
\hline
\textbf{\# Shots}         & \textbf{Model}  & \textbf{R1}                           & \textbf{R2}                           & \textbf{RL}                           & \textbf{RG}                           & \textbf{BertS}                        & \textbf{DiscoS}                       & \textbf{MDSummaC}                                          & \textbf{N-gram Novelty}               \\ \hline
\multicolumn{1}{r|}{0}    & Pegasus-X       & 30.4                                  & 5.6                                   & 15.2                                  & 13.6                                  & 57.1                                  & 91.4                                  & 18.0                                                       & 6.1                                   \\
                          & Primera         & 32.0                                  & 6.6                                   & 16.3                                  & 15.0                                  & 58.0                                  & 83.6                                  & 17.2                                                       & 3.3                                   \\
                          & QAmden          & 25.3                                  & 4.1                                   & 14.1                                  & 11.3                                  & 51.6                                  & 72.7                                  & 6.5                                                        & 29.8                                  \\
                          & Centrum         & 32.8                                  & 7.2                                   & 17.3                                  & 15.8                                  & 58.2                                  & \textbf{92.4}                         & 18.6                                                       & 18.9                                  \\
                          & PELMS           & \cellcolor[HTML]{B7E1CD}\textbf{33.8} & \cellcolor[HTML]{B7E1CD}\textbf{7.9}  & \cellcolor[HTML]{B7E1CD}\textbf{17.7} & \cellcolor[HTML]{B7E1CD}\textbf{16.7} & \cellcolor[HTML]{B7E1CD}\textbf{59.2} & 92.2                                  & \cellcolor[HTML]{B7E1CD}\textbf{19.7}                      & 19.2                                  \\ \hline
\multicolumn{1}{r|}{16}   & Pegasus-X (5\%) & 26.0                                  & 4.5                                   & 14.0                                  & 11.6                                  & 56.8                                  & 83.9                                  & 18.0                                                       & 5.1                                   \\
                          & Primera (5\%)   & 31.1                                  & 6.4                                   & 16.2                                  & 14.6                                  & 58.2                                  & 86.7                                  & 20.2                                                       & 2.3                                   \\
                          & QAmden (5\%)    & 19.0                                  & 3.1                                   & 12.1                                  & 8.8                                   & 53.1                                  & 68.2                                  & 17.4                                                       & 17.4                                  \\
                          & Centrum (5\%)   & 32.6                                  & 7.3                                   & 17.2                                  & 15.8                                  & 58.1                                  & \textbf{92.3}                         & 18.7                                                       & \textbf{18.7}                         \\
                          & PELMS (5\%)     & \cellcolor[HTML]{B1CBF2}\textbf{34.0} & \cellcolor[HTML]{B1CBF2}\textbf{8.1}  & \cellcolor[HTML]{B1CBF2}\textbf{18.1} & \cellcolor[HTML]{B1CBF2}\textbf{17.0} & \cellcolor[HTML]{B1CBF2}\textbf{60.0} & \cellcolor[HTML]{B1CBF2}\textbf{92.3} & \cellcolor[HTML]{B7E1CD}\textbf{20.4}                      & 18.1                                  \\ \cline{2-10} 
                          & Pegasus-X       & 31.9                                  & 6.9                                   & 17.4                                  & 15.5                                  & 60.5                                  & 89.1                                  & 13.0                                                       & \textbf{34.1}                         \\
                          & Primera         & 34.5                                  & 8.1                                   & 18.7                                  & 17.2                                  & 61.1                                  & 92.4                                  & 18.5                                                       & 24.0                                  \\
                          & QAmden          & 19.8                                  & 3.3                                   & 12.6                                  & 9.3                                   & 53.3                                  & 68.4                                  & 17.4                                                       & 17.0                                  \\
                          & Centrum         & 33.0                                  & 7.4                                   & 17.4                                  & 16.0                                  & 58.4                                  & \textbf{92.7}                         & \textbf{19.2}                                              & 16.2                                  \\
                          & PELMS           & \cellcolor[HTML]{B7E1CD}\textbf{36.0} & \cellcolor[HTML]{B7E1CD}\textbf{9.0}  & \cellcolor[HTML]{B7E1CD}\textbf{19.5} & \cellcolor[HTML]{B7E1CD}\textbf{18.4} & \cellcolor[HTML]{B7E1CD}\textbf{61.6} & 92.6                                  & 16.5                                                       & 30.1                                  \\ \hline
\multicolumn{1}{r|}{64}   & Pegasus-X (5\%) & 29.3                                  & 6.0                                   & 16.1                                  & 14.1                                  & 58.5                                  & 86.5                                  & 17.6                                                       & 11.4                                  \\
                          & Primera (5\%)   & 32.7                                  & 7.4                                   & 17.3                                  & 15.9                                  & 59.5                                  & 90.2                                  & \textbf{21.0}                                              & 7.6                                   \\
                          & QAmden (5\%)    & 19.5                                  & 3.2                                   & 12.4                                  & 9.2                                   & 53.3                                  & 68.4                                  & 17.3                                                       & 17.0                                  \\
                          & Centrum (5\%)   & 32.7                                  & 7.4                                   & 17.3                                  & 15.9                                  & 58.4                                  & \textbf{92.6}                         & 19.1                                                       & 16.7                                  \\
                          & PELMS (5\%)     & \cellcolor[HTML]{B1CBF2}\textbf{35.0} & \cellcolor[HTML]{B1CBF2}\textbf{8.8}  & \cellcolor[HTML]{B1CBF2}\textbf{19.0} & \cellcolor[HTML]{B1CBF2}\textbf{17.9} & \cellcolor[HTML]{B1CBF2}\textbf{60.9} & 92.5                                  & 20.9                                                       & \cellcolor[HTML]{B1CBF2}\textbf{18.4} \\ \hline
                          & Pegasus-X       & 35.4                                  & 8.6                                   & 19.4                                  & 18.0                                  & 62.2                                  & 92.0                                  & 10.2                                                       & \textbf{46.1}                         \\
                          & Primera         & 35.8                                  & 8.8                                   & 19.3                                  & 18.2                                  & 61.8                                  & \textbf{93.0}                         & 16.2                                                       & 31.0                                  \\
                          & QAmden          & 29.6                                  & 7.1                                   & 17.3                                  & 15.3                                  & 59.0                                  & 83.2                                  & 17.6                                                       & 20.4                                  \\
                          & Centrum         & 33.5                                  & 7.7                                   & 17.8                                  & 16.4                                  & 59.3                                  & 92.7                                  & \textbf{20.0}                                              & 12.2                                  \\
                          & PELMS           & \cellcolor[HTML]{B7E1CD}\textbf{36.2} & \cellcolor[HTML]{B7E1CD}\textbf{9.3}  & \cellcolor[HTML]{B7E1CD}\textbf{19.8} & \cellcolor[HTML]{B7E1CD}\textbf{18.7} & \cellcolor[HTML]{B7E1CD}\textbf{62.5} & 92.3                                  & 15.5                                                       & 34.4                                  \\ \hline
\multicolumn{1}{r|}{Full} & Pegasus-X (5\%) & 31.4                                  & 7.3                                   & 16.7                                  & 15.4                                  & 59.2                                  & 89                                    & 15.4                                                       & 18.2                                  \\
                          & Primera (5\%)   & 32.2                                  & 7.5                                   & 17.2                                  & 15.9                                  & 59.5                                  & 88.2                                  & 18.9                                                       & 12.5                                  \\
                          & QAmden (5\%)    & 27.6                                  & 6.2                                   & 15.6                                  & 13.6                                  & 57.4                                  & 80.8                                  & 16.9                                                       & 17.3                                  \\
                          & Centrum (5\%)   & 33                                    & 7.8                                   & 17.7                                  & 16.4                                  & 59                                    & 92                                    & 18.4                                                       & \textbf{21.5}                         \\
                          & PELMS (5\%)     & \cellcolor[HTML]{B1CBF2}\textbf{34.5} & \cellcolor[HTML]{B1CBF2}\textbf{8.4}  & \cellcolor[HTML]{B1CBF2}\textbf{18.4} & \cellcolor[HTML]{B1CBF2}\textbf{17.4} & \cellcolor[HTML]{B1CBF2}\textbf{60.3} & \cellcolor[HTML]{B1CBF2}\textbf{92.1} & \cellcolor[HTML]{B7E1CD}\textbf{19.8}                      & 19.4                                  \\ \cline{2-10} 
                          & Pegasus-X       & 36.6                                  & 9.9                                   & 20.1                                  & 19.2                                  & 62.4                                  & 92.1                                  & 13.0                                                       & \textbf{31.7}                         \\
                          & Primera         & 37.1                                  & 10                                    & 20.4                                  & 19.4                                  & 62.6                                  & \textbf{93.3}                         & 16.6                                                       & 30.7                                  \\
                          & QAmden          & 36.6                                  & 9.9                                   & 20.2                                  & 19.2                                  & 62.2                                  & 91.9                                  & 16.1                                                       & 28.4                                  \\
                          & Centrum         & 35                                    & 9.0                                   & 19.0                                  & 18.0                                  & 60.7                                  & 92.8                                  & 16.7                                                       & 23.4                                  \\ \cline{9-9}
                          & PELMS           & \cellcolor[HTML]{B7E1CD}\textbf{37.6} & \cellcolor[HTML]{B7E1CD}\textbf{10.5} & \cellcolor[HTML]{B7E1CD}\textbf{20.8} & \cellcolor[HTML]{B7E1CD}\textbf{20.0} & \cellcolor[HTML]{B7E1CD}\textbf{63.3} & \multicolumn{1}{c|}{93.0}             & \multicolumn{1}{c|}{\cellcolor[HTML]{B1CBF2}\textbf{17.0}} & 31.0                                  \\ \hline
\multicolumn{1}{r|}{LLM}  & GPT-3.5-Long    & 34.6                                  & 7.6                                   & 18.1                                  & 16.8                                  & 61.2                                  & 91.5                                  & 13.6                                                       & 50.9                                  \\
                          & GPT-4           & 35.2                                  & 7.9                                   & 18.4                                  & 17.1                                  & 61.8                                  & 92.0                                  & 11.6                                                       & 52.1                                  \\ \hline
\end{tabular}

%% file: tables/model_ablation.tex
\begin{tabular}{llc|ccccc}
\hline
\textbf{Architecture} & \textbf{Technique} & \textbf{w/ MultiPT} & \textbf{RG}   & \textbf{BertScore} & \textbf{DiscoScore} & \textbf{MDSummaC} & \textbf{N-gram Novelty} \\ \hline
LED                   & Primera            & No                  & 15.0          & 58.0               & 91.4                & 18.0              & 3.3                     \\
LED                   & Primera            & Yes                 & 15.4          & 58.4               & 91.5                & \textbf{20.7}     & 4.6                     \\
LED                   & PELMS              & Yes                 & \textbf{16.7} & \textbf{59.2}      & \textbf{92.2}       & 19.7              & \textbf{19.2}           \\ \hline
Pegasus-X             & Pegasus-X           & No                  & 13.6          & 57.1               & 91.4                & \textbf{18.0}     & 6.1                     \\
Pegasus-X             & PELMS              & Yes                 & \textbf{17.3} & \textbf{59.8}      & \textbf{92.8}       & 17.6              & \textbf{22.6}           \\ \hline
\end{tabular}

%% file: tables/human_eval.tex
\begin{tabular}{l|cccc}
\hline
          & \multicolumn{4}{c}{MetaTomatoes (25 Examples)}                                    \\
          & Gram. $\downarrow$        & Ref. $\downarrow$         & \multicolumn{1}{c|}{Str. \& Coher $\downarrow$} & Faithf. $\uparrow$     \\ \hline
Pegasus-X & 1.9          & 1.88          & \multicolumn{1}{c|}{2.02}          & 65.9          \\
Primera   & 1.85         & 1.72          & \multicolumn{1}{c|}{1.85}          & 58.4          \\
PELMS     & \textbf{1.8} & \textbf{1.63} & \multicolumn{1}{c|}{\textbf{1.58}} & \textbf{78.3} \\ \hline
\end{tabular}

%% file: tables/eval_datasets.tex
\begin{tabular}{lcrrrrr}

& \multicolumn{1}{l}{} & \multicolumn{1}{l}{}                    & \multicolumn{1}{l}{}                    & \multicolumn{1}{l}{}                  & \multicolumn{1}{l}{}                    & \multicolumn{1}{l}{}                        \\ \hline
\textbf{Evaluation Datasets}           & \textbf{Domain}      & \multicolumn{1}{c}{\textbf{\#Clusters}} & \multicolumn{1}{c}{\textbf{\#Docs / C}} & \multicolumn{1}{c}{\textbf{Doc\_len}} & \multicolumn{1}{c}{\textbf{Input\_len}} & \multicolumn{1}{c}{\textbf{Summ. Length}}   \\ \hline
MultiNews                             & News                 & 56,000                                  & 2.8                                     & 640.4                                 & 1793                                    & 217                                         \\
Multi-XScience                        & Academic Literature  & 40,000                                  & 4.4                                     & 160                                   & 700                                   &   105                                         \\
Amazon                                & Product Reviews      & 180                                     & 8                                       & 49.7                                  & 397                                   & 50.3                                        \\
Yelp                                  & Business Reviews     & 300                                     & 8                                       & 49.8                                  & 398.4                                   & 52.3                                        \\
DUC2004                               & News                 & 50                                      & 10                                      & 588.2                                 & 5882                                    & 115                                         \\
MetaTomatoes                          & Movie Meta-Reviews   & 1,497                                   & 84.3                                    & 23.2                                  & 1956                                  & 142                                       \\ \hline

\end{tabular}

%% file: tables/control_results.tex
\begin{tabular}{l|l|cccccccc}
\hline
\textbf{Dataset}  & \textbf{Best Prefix} & \textbf{R1}   & \textbf{R2}  & \textbf{RL}   & \textbf{RG}   & \textbf{BertS} & \textbf{DiscoS} & \textbf{MDSumC} & \textbf{N-gram Novelty} \\ \hline
MultiNews         & long                 & 43.4          & 14.1         & 20.6          & 23.3          & 61.4           & 95.2            & 38.1            & 9.7                     \\
Multi-X-Science   & none                 & 29.7          & 5.1          & 15.6          & 13.3          & 56.4           & 90.8            & 30.9            & 5.5                     \\
Amazon            & medium-long          & 35.4          & 8.7          & 21.2          & 18.7          & 63.2           & 90.9            & 15.3            & 32.4                    \\
Yelp              & short-medium         & 30.8          & 6.6          & 18.1          & 15.4          & 61.3           & 91.2            & 10.2            & 38.2                    \\
Duc2004           & medium               & 36.1          & 8.4          & 18.0          & 17.6          & 58.3           & 93.7            & 14.8            & 7.3                     \\
MetaTomatoes      & medium-long          & 33.8          & 7.1          & 16.3          & 15.7          & 55.0           & 92.0            & 6.8             & 43.6                    \\ \hline
Controlled Avg.   &                      & \textbf{34.9} & \textbf{8.3} & \textbf{18.3} & \textbf{17.3} & \textbf{59.3}  & \textbf{92.3}   & 19.4            & \textbf{22.8}           \\
Uncontrolled Avg. &                      & 33.8          & 7.9          & 17.7          & 16.7          & 59.2           & 92.2            & \textbf{19.7}   & 19.2                    \\ \hline
\end{tabular}

%% file: tables/model_summaries.tex
\begingroup
\renewcommand{\arraystretch}{1.5}
\setlength{\tabcolsep}{10pt}
\ttfamily
\begin{tabularx}{\textwidth}{XXXX}
\textbf{Ground Truth} & \textbf{Pegasus-X} & \textbf{Primera} & \textbf{PELMS} \\ \hline

 What do you do with an ex-pope? If you happen to know, call Rome, because the Church isn't sure yet. Many fear Benedict will become an implicit rival to the new pope, despite his apparent desire to keep a low profile. He told Roman diocese priests yesterday that he would be "withdrawing into prayer," Reuters reports. "I will always be close to all of you," he said, "even if I remain hidden from the world." The "close to you" bit is somewhat literal; Benedict has announced he'll be living in the Vatican. That's caused some controversy, the Wall Street Journal reports, but one senior papal adviser says it's "better to have him here than somewhere else, where he could become another center of power." Benedict is likely to exert influence over the next pope by picking a new head of the embattled Vatican Bank, and he'll keep writing as a respected theologian. The Church also hasn't decided what vestments he'll wear, whether he'll give audiences, or even what to address him ‘a decision that has "juridical" ramifications’, a spokesman said. One thing's for sure: He won't be infallible anymore.
     &      
 
VATICAN CITY - The looming retirement of Pope Benedict XVI, who will soon become the first living ex-pontiff in some six centuries, has created a quandary for the Roman Catholic Church: What to make of a former pope who will continue to live under the roof of a new one? Speaking unscripted to thousands of priests from the diocese of Rome, in what turned out to be a farewell address in his capacity as bishop of the Italian capital, Benedict outlined a cloistered life ahead, once he steps down in two weeks time: "Even if I am withdrawing into prayer, I will always been close to all of you and I am sure that you will be close to me, even if I remain hidden to the world," he said. After February 28, when he becomes the first pontiff to resign instead of ruling for life, Benedict will first go to the papal summer retreat at Castel Gandolfo, south of Rome and then move permanently into the four-storey Mater Ecclesiae convent.
     &   

VATICAN CITY - The looming retirement of Pope Benedict XVI, who will soon become the first living ex-pontiff in some six centuries, has created a quandary for the Roman Catholic Church: What to make of a former pope who will continue to live under the roof of a new one? Wednesday provided two examples of how the former officeholder could cast a shadow over his successor's work. Pope Benedict was likely to take the sensitive step of appointing a new chief of the Vatican bank, Vatican spokesman Rev. Federico Lombardi said - a move that would leave the next pontiff with a just-picked custodian to a secretive bank that, under Pope Benedict, has denied allegations of money-laundering. The pope underscored that Wednesday. "I have done this in full freedom for the good of the church," he told a general audience, one of his last. Later at Mass, the pontiff received a long and emotional ovation from the hundreds of rank-and-file faithful who packed the pews, including dozens of cardinals who removed their miters in a sign of respect to the outgoing pontiff. One cleric wiped away tears.    &

Pope Benedict XVI's decision to resign is fueling an unusual discussion within the Vatican: What status he'll hold once he retires? The Vatican, which is navigating uncharted waters since his shock announcement, said experts have still not decided what his title will be or whether he will wear the white of a pope, the red of a cardinal or the black of an ordinary priest. Pope Benedict also made his first public appearances in the wake of his resignation announcement, issuing a strong warning to the cardinals who will elect his successor, decrying the internal rivalries among senior churchmen that have plagued his eight-year papacy. The rivalries burst into public view a year ago when the Vatican confirmed the authenticity of letters to the pope, leaked to the Italian media, detailing the ouster of a top Vatican official who tried to blow the whistle on wasteful spending in the world's smallest state. Father Lombardi said he didn't think the pope consulted the College of Cardinals on his decision to stay inside the Vatican. Added a senior papal adviser: “It’s better to have him here than somewhere else, where he could become another center of power.”   \\ \hline

 \textbf{Centrum} & \textbf{QAMDen} & \textbf{GPT-3.5-long} & \textbf{GPT-4} \\ \hline

VATICAN CITY (Reuters) - Pope Benedict XVI’s decision to resign is fueling an unusual discussion within the Vatican: What status he’ll hold once he retires? FILE PHOTO: Pope Benedict Benedict XVI attends Ash Wednesday mass at the Vatican February 13, 2013. Thousands of people are expected to gather in the Vatican for Pope Benedict's Ash Wednesday Mass, which is expected to be his last before leaving office at the end of February. VATICAN CITY - Pope Benedict will see out his life in prayer, ‘hidden from the world’, he said on Thursday in his first personal comment on his plans since he stunned Roman Catholics by announcing his retirement. His remarks, in a voice that was hoarse at times, followed Monday's resignation notice which spoke of ‘a life dedicated to prayer’. The Vatican has said the 85-year-old German will live within its walls. His seclusion may allay concern that the first living former pope in centuries might trouble Church unity. Speaking unscripted to thousands of priests from the diocese of Rome, in what turned out to be a farewell address in his capacity as bishop of the Italian capital, Benedict outlined a cloistered life ahead, once he steps down in two weeks time.
 &

   a cloistered life ahead, becauese speaking unscripted to thousands of priests from the diocese of Rome, in what turned out to be a farewell address in his capacity as bishop of the Italian capital, Benedict outlined a close relationship with his flock, once he steps down in two weeks time - - Advertisement - \"The face of the church is at times disfigured by the sins against the unity of the Church, who will never stop guiding it and I am sure that you will be close to all of you, even if I remain hidden to the world,\" he said, \"Even if I am withdrawing into prayer, I will always be close, and I'm sure that I, myself, will always, I'm very sorry that I remain, I hope that I'll always be very close to you, I'll see you, but I'll be very very, very close, very very very close and I hope you'll be close.
. &

   Pope Benedict XVI's decision to resign has sparked discussion about his status once he retires. There are concerns about whether he will become a rival to his successor, as he plans to live inside the Vatican. Pope Benedict has made it clear that he will not interfere with his successor's work and will give up all formal powers of his office. However, there are still questions about what title to use for him and whether he will hold audiences with potential admirers. After his resignation, Pope Benedict will live a secluded life of prayer and will move into the Mater Ecclesiae convent within the Vatican. The Vatican has not yet decided on his title or what he will wear. The election of the new pope will take place in a secret conclave between March 15 and 20.
 &

Pope Benedict XVI's decision to resign has sparked discussions within the Vatican about his status post-retirement. This is the first time in six centuries that a living ex-pontiff will coexist with a new Pope, raising questions about potential conflicts. Pope Benedict has made plans to live inside the Vatican after his resignation, which some analysts believe could unwittingly make him a rival to his successor. However, he has stated that he has no intention of interfering with his successor's work and will give up all formal powers of his office. The Pope will continue to live inside Vatican City, where he will write and pray. His future home will be a renovated convent within Vatican walls. The Vatican has stated that he will not influence the election of his successor. However, his title post-resignation and his attire are still under consideration. The Pope has stated that he will spend his time in prayer, "hidden from the world”.
\end{tabularx}
\endgroup

%% file: tables/_llm_full.tex
\begin{tabular}{ll|cccccccc}
\hline
\textbf{Dataset} & \textbf{Model} & \textbf{R1} & \textbf{R2} & \textbf{RL} & \textbf{RG} & \textbf{BertS} & \textbf{DiscoS} & \textbf{MDSummaC} & \textbf{N-gram Novelty} \\ \hline
MultiNews        & GPT-3.5-Long   & 40.6        & 12.2        & 20.2        & 21.6        & 63.0           & 94.0            & 27.4              & 38.9                    \\
                 & GPT-4          & 41.3        & 12.5        & 20.3        & 21.9        & 63.4           & 94.5            & 26.3              & 37.5                    \\ \hline
Multi-XScience   & GPT-3.5-Long   & 30.7        & 5.2         & 15.8        & 13.7        & 56.5           & 91.1            & 20.6              & 40.6                    \\
                 & GPT-4          & 30.8        & 5.1         & 15.9        & 13.6        & 56.7           & 91.4            & 16.9              & 42.9                    \\ \hline
Amazon           & GPT-3.5-Long   & 31.3        & 5.7         & 17.9        & 14.7        & 62.4           & 90.6            & 6.6               & 70.1                    \\
                 & GPT-4          & 33.2        & 6.9         & 19.1        & 16.3        & 64.6           & 91.9            & 5.2               & 71.3                    \\ \hline
Yelp             & GPT-3.5-Long   & 33.6        & 6.6         & 19.3        & 16.2        & 64.0           & 91.9            & 6.9               & 65.2                    \\
                 & GPT-4          & 32.7        & 6.0         & 18.5        & 15.4        & 64.3           & 91.5            & 4.6               & 70.2                    \\ \hline
DUC2004          & GPT-3.5-Long   & 40.0        & 10.4        & 20.2        & 20.3        & 63.3           & 95.4            & 14.3              & 34.4                    \\
                 & GPT-4          & 40.8        & 11.2        & 21.2        & 21.3        & 64.0           & 95.4            & 12.1              & 34.2                    \\ \hline
MetaTomatoes     & GPT-3.5-Long   & 31.2        & 5.7         & 15.2        & 14.0        & 57.9           & 85.9            & 6.0               & 55.9                    \\
                 & GPT-4          & 32.2        & 5.7         & 15.4        & 14.1        & 57.8           & 87.5            & 4.5               & 56.2                    \\ \hline
Overall          & GPT-3.5-Long   & 35.4        & 7.8         & 18.1        & 17.0        & 61.5           & 91.1            & 8.3               & 50.2                    \\
                 & GPT-4          & 35.7        & 8.0         & 18.9        & 17.4        & 62.7           & 92.0            & 8.8               & 52.0                    \\ \hline
\end{tabular}

%% file: tables/16_full_parameter.tex
\begin{tabular}{ll|llllllll}
\hline
\textbf{Dataset} & \textbf{Model} & \textbf{R1}                                                    & \textbf{R2}                                                   & \textbf{RL}                                                    & \textbf{RG}                                                    & \textbf{BertS}                                                 & \textbf{DiscoS}                                      & \textbf{MDSummaC}                                             & \textbf{N-gram Novelty}                                        \\ \hline
MultiNews        & Pegasus-X      & 35.5 (0.2)                                                     & 11.3 (0.5)                                                    & 18.0 (0.4)                                                     & 19.3 (0.5)                                                     & 60.6 (0.5)                                                     & 85.0 (0.8)                                           & 28.0 (2.8)                                                    & 17.3 (4.1)                                                     \\
                 & Primera        & 44.2 (0.3)                                                     & 15.1 (0.4)                                                    & 21.8 (0.4)                                                     & 24.4 (0.4)                                                     & 63.0 (0.3)                                                     & 95.6 (0.2)                                           & 36.1 (3.7)                                                    & 11.4 (3.7)                                                     \\
\hline
\end{tabular}

%% file: tables/16_adapter.tex
\begin{tabular}{ll|llllllll}
\hline
\textbf{Dataset} & \textbf{Model} & \textbf{R1}                                                    & \textbf{R2}                                                   & \textbf{RL}                                                    & \textbf{RG}                                                    & \textbf{BertS}                                                 & \textbf{DiscoS}                                                & \textbf{MDSummaC}                                             & \textbf{N-gram Novelty}                                        \\ \hline
MultiNews        & Pegasus-X      & 25.1 (0.4)                                                     & 7.2 (0.2)                                                     & 13.6 (0.2)                                                     & 13.5 (0.3)                                                     & 56.9 (0.1)                                                     & 71.0 (0.5)                                                     & 40.0 (0.3)                                                    & 1.7 (0.3)                                                      \\
                 & Primera        & 41.1 (0.6)                                                     & 12.8 (0.4)                                                    & 19.7 (0.4)                                                     & 21.8 (0.5)                                                     & 61.0 (0.3)                                                     & 93.5 (1.1)                                                     & 43.3 (0.4)                                                    & 1.8 (0.2)                                                      \\
\hline
\end{tabular}

%% file: tables/64_full_parameter.tex
\begin{tabular}{ll|llllllll}
\hline
\textbf{Dataset} & \textbf{Model} & \textbf{R1}                                                    & \textbf{R2}                                                   & \textbf{RL}                                          & \textbf{RG}                                                    & \textbf{BertS}                                       & \textbf{DiscoS}                                      & \textbf{MDSummaC} & \textbf{N-gram Novelty} \\ \hline
MultiNews        & Pegasus-X      & 42.0 (0.5)                                                     & 14.1 (0.2)                                                    & 20.8 (0.3)                                           & 23.1 (0.3)                                                     & 62.8 (0.3)                                           & 93.5 (0.5)                                           & 23.2 (3.5)        & 26.5 (4.6)              \\
                 & Primera        & 45.0 (0.4)                                                     & 15.9 (0.5)                                                    & 22.3 (0.5)                                           & 25.2 (0.4)                                                     & 63.7 (0.3)                                           & 96.0 (0.2)                                           & 33.2 (3.8)        & 17.0 (3.9)              \\
                 & QAmden         & 29.2 (1.0)                                                     & 9.2 (0.4)                                                     & 16.0 (0.4)                                           & 16.3 (0.5)                                                     & 57.2 (0.4)                                           & 76.1 (1.4)                                           & 36.3 (0.4)        & 8.6 (0.2)               \\
\hline
\end{tabular}

%% file: tables/64_adapter.tex
\begin{tabular}{ll|llllllll}
\hline
\textbf{Dataset} & \textbf{Model} & \textbf{R1}                                                    & \textbf{R2}                                                   & \textbf{RL}                                                    & \textbf{RG}                                                    & \textbf{BertS}                                                 & \textbf{DiscoS}                                                & \textbf{MDSummaC}                                              & \textbf{N-gram Novelty}                                        \\ \hline
MultiNews        & Pegasus-X      & 26.2 (0.6)                                                     & 7.6 (0.4)                                                     & 14.1 (0.4)                                                     & 14.1 (0.5)                                                     & 57.1 (0.2)                                                     & 73.1 (0.8)                                                     & 39.1 (0.5)                                                     & 2.6 (0.3)                                                      \\
                 & Primera        & 42.2 (0.5)                                                     & 13.8 (0.5)                                                    & 20.6 (0.5)                                                     & 22.9 (0.5)                                                     & 61.8 (0.5)                                                     & 94.8 (0.1)                                                     & 43.9 (0.2)                                                     & 2.2 (0.3)                                                      \\
\hline
\end{tabular}

%% file: tables/full_shot_full_parameter.tex
\begin{tabular}{ll|llllllll}
\hline
\textbf{Dataset} & \textbf{Model} & \textbf{R1}                                                    & \textbf{R2}                                                    & \textbf{RL}                                                    & \textbf{RG}                                                    & \textbf{BertS}                                                 & \textbf{DiscoS}                                                & \textbf{MDSummaC}                                    & \textbf{N-gram Novelty}                                        \\ \hline
MultiNews        & Pegasus-X      & 47.4 (0.4)                                                     & 18.8 (0.4)                                                     & 24.3 (0.2)                                                     & 27.9 (0.3)                                                     & 65.4 (0.4)                                                     & 95.9 (0.4)                                                     & 26.8 (1.4)                                           & 18.4 (1.4)                                                     \\
                 & Primera        & 47.6 (0.3)                                                     & 18.4 (0.6)                                                     & 24.2 (0.2)                                                     & 27.7 (0.4)                                                     & 65.2 (0.4)                                                     & 96.5 (0.7)                                                     & 34.6 (1.1)                                           & 16.6 (0.8)                                                     \\
\hline
\end{tabular}

%% file: tables/full_shot_adapter.tex
\begin{tabular}{ll|llllllll}
\hline
\textbf{Dataset} & \textbf{Model} & \textbf{R1}                                                    & \textbf{R2}                                                   & \textbf{RL}                                                    & \textbf{RG}                                                    & \textbf{BertS}                                                 & \textbf{DiscoS}                                                & \textbf{MDSummaC}                                              & \textbf{N-gram Novelty}                                        \\ \hline
MultiNews        & Pegasus-X      & 44.6 (0.4)                                                     & 16.3 (0.4)                                                    & 22.6 (0.3)                                                     & 25.4 (0.3)                                                     & 63.8 (0.5)                                                     & 95.8 (0.3)                                                     & 35.6 (0.4)                                                     & 10.4 (2.1)                                                     \\
                 & Primera        & 44.4 (0.4)                                                     & 15.4 (0.3)                                                    & 22.0 (0.5)                                                     & 24.7 (0.4)                                                     & 63.1 (0.3)                                                     & 96.1 (0.2)                                                     & 42.2 (0.6)                                                     & 6.0 (0.9)                                                      \\
\hline
\end{tabular}

%% file: acl_latex.bbl
\begin{thebibliography}{45}
\expandafter\ifx\csname natexlab\endcsname\relax\def\natexlab#1{#1}\fi

\bibitem[{Beltagy et~al.(2020)Beltagy, Peters, and Cohan}]{longformer_led}
Iz~Beltagy, Matthew~E. Peters, and Arman Cohan. 2020.
\newblock \href {http://arxiv.org/abs/2004.05150} {Longformer: The
  long-document transformer}.
\newblock \emph{CoRR}, abs/2004.05150.

\bibitem[{Bra{\v{z}}inskas et~al.(2020)Bra{\v{z}}inskas, Lapata, and
  Titov}]{amazon_yelp_bravzinskas2020few}
Arthur Bra{\v{z}}inskas, Mirella Lapata, and Ivan Titov. 2020.
\newblock Few-shot learning for opinion summarization.
\newblock \emph{arXiv preprint arXiv:2004.14884}.

\bibitem[{Brazinskas et~al.(2022)Brazinskas, Nallapati, Bansal, and
  Dreyer}]{brazinskas-etal-2022-efficient}
Arthur Brazinskas, Ramesh Nallapati, Mohit Bansal, and Markus Dreyer. 2022.
\newblock \href {https://doi.org/10.18653/v1/2022.findings-naacl.113}
  {Efficient few-shot fine-tuning for opinion summarization}.
\newblock In \emph{Findings of the Association for Computational Linguistics:
  NAACL 2022}, pages 1509--1523, Seattle, United States. Association for
  Computational Linguistics.

\bibitem[{Caciularu et~al.(2021)Caciularu, Cohan, Beltagy, Peters, Cattan, and
  Dagan}]{cdlm-cross-document-language-modeling}
Avi Caciularu, Arman Cohan, Iz~Beltagy, Matthew~E. Peters, Arie Cattan, and Ido
  Dagan. 2021.
\newblock \href {http://arxiv.org/abs/2101.00406} {Cross-document language
  modeling}.
\newblock \emph{CoRR}, abs/2101.00406.

\bibitem[{Caciularu et~al.(2023)Caciularu, Peters, Goldberger, Dagan, and
  Cohan}]{caciularu-etal-2023-peek-qamden}
Avi Caciularu, Matthew Peters, Jacob Goldberger, Ido Dagan, and Arman Cohan.
  2023.
\newblock \href {https://doi.org/10.18653/v1/2023.acl-long.110} {Peek across:
  Improving multi-document modeling via cross-document question-answering}.
\newblock In \emph{Proceedings of the 61st Annual Meeting of the Association
  for Computational Linguistics (Volume 1: Long Papers)}, pages 1970--1989,
  Toronto, Canada. Association for Computational Linguistics.

\bibitem[{Cao and Wang(2021)}]{cao2021cliff}
Shuyang Cao and Lu~Wang. 2021.
\newblock \href {http://arxiv.org/abs/2109.09209} {Cliff: Contrastive learning
  for improving faithfulness and factuality in abstractive summarization}.

\bibitem[{Christensen et~al.(2013)Christensen, {Mausam}, Soderland, and
  Etzioni}]{christensen-etal-2013-towards_coherence}
Janara Christensen, {Mausam}, Stephen Soderland, and Oren Etzioni. 2013.
\newblock \href {https://aclanthology.org/N13-1136} {Towards coherent
  multi-document summarization}.
\newblock In \emph{Proceedings of the 2013 Conference of the North {A}merican
  Chapter of the Association for Computational Linguistics: Human Language
  Technologies}, pages 1163--1173, Atlanta, Georgia. Association for
  Computational Linguistics.

\bibitem[{Chung et~al.(2022)Chung, Hou, Longpre, Zoph, Tay, Fedus, Li, Wang,
  Dehghani, Brahma et~al.}]{flan-t5}
Hyung~Won Chung, Le~Hou, Shayne Longpre, Barret Zoph, Yi~Tay, William Fedus,
  Eric Li, Xuezhi Wang, Mostafa Dehghani, Siddhartha Brahma, et~al. 2022.
\newblock Scaling instruction-finetuned language models.
\newblock \emph{arXiv preprint arXiv:2210.11416}.

\bibitem[{Dang(2005)}]{duc2004_dang2005overview}
Hoa~Trang Dang. 2005.
\newblock Overview of duc 2005.
\newblock In \emph{Proceedings of the document understanding conference},
  volume 2005, pages 1--12. Citeseer.

\bibitem[{DeYoung et~al.(2023)DeYoung, Martinez, Marshall, and
  Wallace}]{deyoung2023multi_synthesis}
Jay DeYoung, Stephanie~C Martinez, Iain~J Marshall, and Byron~C Wallace. 2023.
\newblock Do multi-document summarization models synthesize?
\newblock \emph{arXiv preprint arXiv:2301.13844}.

\bibitem[{Emerson(2013)}]{borda_count}
Peter Emerson. 2013.
\newblock The original borda count and partial voting.
\newblock \emph{Social Choice and Welfare}, 40(2):353--358.

\bibitem[{Fabbri et~al.(2019)Fabbri, Li, She, Li, and
  Radev}]{fabbri2019multinews}
Alexander~R Fabbri, Irene Li, Tianwei She, Suyi Li, and Dragomir~R Radev. 2019.
\newblock Multi-news: A large-scale multi-document summarization dataset and
  abstractive hierarchical model.
\newblock \emph{arXiv preprint arXiv:1906.01749}.

\bibitem[{Gu et~al.(2020)Gu, Mao, Han, Liu, Yu, Wu, Yu, Finnie, Zhai, and
  Zukoski}]{gu2020generating_newshead}
Xiaotao Gu, Yuning Mao, Jiawei Han, Jialu Liu, Hongkun Yu, You Wu, Cong Yu,
  Daniel Finnie, Jiaqi Zhai, and Nicholas Zukoski. 2020.
\newblock \href {http://arxiv.org/abs/2001.09386} {Generating representative
  headlines for news stories}.

\bibitem[{Guo et~al.(2022)Guo, Ainslie, Uthus, Ontanon, Ni, Sung, and
  Yang}]{guo2022longt5}
Mandy Guo, Joshua Ainslie, David Uthus, Santiago Ontanon, Jianmo Ni, Yun-Hsuan
  Sung, and Yinfei Yang. 2022.
\newblock \href {http://arxiv.org/abs/2112.07916} {Longt5: Efficient
  text-to-text transformer for long sequences}.

\bibitem[{Hendrickx et~al.(2009)Hendrickx, Daelemans, Marsi, and
  Krahmer}]{hendrickx2009reducing_redundancy}
Iris Hendrickx, Walter Daelemans, Erwin Marsi, and Emiel Krahmer. 2009.
\newblock Reducing redundancy in multi-document summarization using lexical
  semantic similarity.
\newblock In \emph{Proceedings of the 2009 Workshop on Language Generation and
  Summarisation (UCNLG+ Sum 2009)}, pages 63--66.

\bibitem[{Houlsby et~al.(2019)Houlsby, Giurgiu, Jastrzebski, Morrone,
  De~Laroussilhe, Gesmundo, Attariyan, and
  Gelly}]{pmlr-v97-houlsby19a_adapters}
Neil Houlsby, Andrei Giurgiu, Stanislaw Jastrzebski, Bruna Morrone, Quentin
  De~Laroussilhe, Andrea Gesmundo, Mona Attariyan, and Sylvain Gelly. 2019.
\newblock \href {https://proceedings.mlr.press/v97/houlsby19a.html}
  {Parameter-efficient transfer learning for {NLP}}.
\newblock In \emph{Proceedings of the 36th International Conference on Machine
  Learning}, volume~97 of \emph{Proceedings of Machine Learning Research},
  pages 2790--2799. PMLR.

\bibitem[{Kry{\'s}ci{\'n}ski et~al.(2019)Kry{\'s}ci{\'n}ski, McCann, Xiong, and
  Socher}]{kryscinski2019evaluating_factcc}
Wojciech Kry{\'s}ci{\'n}ski, Bryan McCann, Caiming Xiong, and Richard Socher.
  2019.
\newblock Evaluating the factual consistency of abstractive text summarization.
\newblock \emph{arXiv preprint arXiv:1910.12840}.

\bibitem[{Laban et~al.(2021)Laban, Schnabel, Bennett, and
  Hearst}]{laban2021summac}
Philippe Laban, Tobias Schnabel, Paul~N. Bennett, and Marti~A. Hearst. 2021.
\newblock \href {http://arxiv.org/abs/2111.09525} {Summac: Re-visiting
  nli-based models for inconsistency detection in summarization}.

\bibitem[{Ladhak et~al.(2021)Ladhak, Durmus, He, Cardie, and
  McKeown}]{ladhak2021faithful}
Faisal Ladhak, Esin Durmus, He~He, Claire Cardie, and Kathleen McKeown. 2021.
\newblock Faithful or extractive? on mitigating the
  faithfulness-abstractiveness trade-off in abstractive summarization.
\newblock \emph{arXiv preprint arXiv:2108.13684}.

\bibitem[{Lewis et~al.(2019)Lewis, Liu, Goyal, Ghazvininejad, Mohamed, Levy,
  Stoyanov, and Zettlemoyer}]{lewis2019bart}
Mike Lewis, Yinhan Liu, Naman Goyal, Marjan Ghazvininejad, Abdelrahman Mohamed,
  Omer Levy, Ves Stoyanov, and Luke Zettlemoyer. 2019.
\newblock Bart: Denoising sequence-to-sequence pre-training for natural
  language generation, translation, and comprehension.
\newblock \emph{arXiv preprint arXiv:1910.13461}.

\bibitem[{Li et~al.(2022)Li, Peng, He, Galley, Yu, and Gao}]{li2022dionysus}
Yu~Li, Baolin Peng, Pengcheng He, Michel Galley, Zhou Yu, and Jianfeng Gao.
  2022.
\newblock \href {http://arxiv.org/abs/2212.10018} {Dionysus: A pre-trained
  model for low-resource dialogue summarization}.

\bibitem[{Lin(2004)}]{lin2004rouge}
Chin-Yew Lin. 2004.
\newblock Rouge: A package for automatic evaluation of summaries.
\newblock In \emph{Text summarization branches out}, pages 74--81.

\bibitem[{Liu et~al.(2018)Liu, Saleh, Pot, Goodrich, Sepassi, Kaiser, and
  Shazeer}]{liu2018generating_wikisumog}
Peter~J. Liu, Mohammad Saleh, Etienne Pot, Ben Goodrich, Ryan Sepassi, Lukasz
  Kaiser, and Noam Shazeer. 2018.
\newblock \href {http://arxiv.org/abs/1801.10198} {Generating wikipedia by
  summarizing long sequences}.

\bibitem[{Liu and Lapata(2019)}]{wikisum40}
Yang Liu and Mirella Lapata. 2019.
\newblock \href {https://doi.org/10.18653/v1/P19-1500} {Hierarchical
  transformers for multi-document summarization}.
\newblock In \emph{Proceedings of the 57th Annual Meeting of the Association
  for Computational Linguistics}, pages 5070--5081, Florence, Italy.
  Association for Computational Linguistics.

\bibitem[{Liu et~al.(2022)Liu, Zhang, Wegsman, Beauchamp, and
  Wang}]{liu-etal-2022-politics-bignews}
Yujian Liu, Xinliang~Frederick Zhang, David Wegsman, Nicholas Beauchamp, and
  Lu~Wang. 2022.
\newblock \href {https://doi.org/10.18653/v1/2022.findings-naacl.101}
  {{POLITICS}: Pretraining with same-story article comparison for ideology
  prediction and stance detection}.
\newblock In \emph{Findings of the Association for Computational Linguistics:
  NAACL 2022}, pages 1354--1374, Seattle, United States. Association for
  Computational Linguistics.

\bibitem[{Lu et~al.(2020)Lu, Dong, and Charlin}]{lu2020multixscience}
Yao Lu, Yue Dong, and Laurent Charlin. 2020.
\newblock Multi-xscience: A large-scale dataset for extreme multi-document
  summarization of scientific articles.
\newblock \emph{arXiv preprint arXiv:2010.14235}.

\bibitem[{Ma et~al.(2022)Ma, Zhang, Guo, Wang, and Sheng}]{survey2022mds}
Congbo Ma, Wei~Emma Zhang, Mingyu Guo, Hu~Wang, and Quan~Z Sheng. 2022.
\newblock Multi-document summarization via deep learning techniques: A survey.
\newblock \emph{ACM Computing Surveys}, 55(5):1--37.

\bibitem[{Nenkova et~al.(2007)Nenkova, Passonneau, and
  McKeown}]{nenkova2007pyramid}
Ani Nenkova, Rebecca Passonneau, and Kathleen McKeown. 2007.
\newblock The pyramid method: Incorporating human content selection variation
  in summarization evaluation.
\newblock \emph{ACM Transactions on Speech and Language Processing (TSLP)},
  4(2):4--es.

\bibitem[{Ni et~al.(2019)Ni, Li, and McAuley}]{ni-etal-2019-amazon}
Jianmo Ni, Jiacheng Li, and Julian McAuley. 2019.
\newblock \href {https://doi.org/10.18653/v1/D19-1018} {Justifying
  recommendations using distantly-labeled reviews and fine-grained aspects}.
\newblock In \emph{Proceedings of the 2019 Conference on Empirical Methods in
  Natural Language Processing and the 9th International Joint Conference on
  Natural Language Processing (EMNLP-IJCNLP)}, pages 188--197, Hong Kong,
  China. Association for Computational Linguistics.

\bibitem[{Otmakhova et~al.(2022)Otmakhova, Truong, Baldwin, Cohn, Verspoor, and
  Lau}]{led-global-attention}
Yulia Otmakhova, Thinh~Hung Truong, Timothy Baldwin, Trevor Cohn, Karin
  Verspoor, and Jey~Han Lau. 2022.
\newblock \href {https://aclanthology.org/2022.sdp-1.21} {{LED} down the rabbit
  hole: exploring the potential of global attention for biomedical
  multi-document summarisation}.
\newblock In \emph{Proceedings of the Third Workshop on Scholarly Document
  Processing}, pages 181--187, Gyeongju, Republic of Korea. Association for
  Computational Linguistics.

\bibitem[{Pasunuru et~al.(2021)Pasunuru, Liu, Bansal, Ravi, and
  Dreyer}]{pasunuru-etal-2021-efficiently}
Ramakanth Pasunuru, Mengwen Liu, Mohit Bansal, Sujith Ravi, and Markus Dreyer.
  2021.
\newblock \href {https://doi.org/10.18653/v1/2021.naacl-main.380} {Efficiently
  summarizing text and graph encodings of multi-document clusters}.
\newblock In \emph{Proceedings of the 2021 Conference of the North American
  Chapter of the Association for Computational Linguistics: Human Language
  Technologies}, pages 4768--4779, Online. Association for Computational
  Linguistics.

\bibitem[{Phang et~al.(2022)Phang, Zhao, and Liu}]{phang2022pegasusx}
Jason Phang, Yao Zhao, and Peter~J. Liu. 2022.
\newblock \href {http://arxiv.org/abs/2208.04347} {Investigating efficiently
  extending transformers for long input summarization}.

\bibitem[{Puduppully et~al.(2023)Puduppully, Jain, Chen, and
  Steedman}]{puduppully-centrum}
Ratish~Surendran Puduppully, Parag Jain, Nancy Chen, and Mark Steedman. 2023.
\newblock \href {https://doi.org/10.18653/v1/2023.acl-short.13} {Multi-document
  summarization with centroid-based pretraining}.
\newblock In \emph{Proceedings of the 61st Annual Meeting of the Association
  for Computational Linguistics (Volume 2: Short Papers)}, pages 128--138,
  Toronto, Canada. Association for Computational Linguistics.

\bibitem[{Radev(2000)}]{radev-2000-common}
Dragomir Radev. 2000.
\newblock \href {https://doi.org/10.3115/1117736.1117745} {A common theory of
  information fusion from multiple text sources step one: Cross-document
  structure}.
\newblock In \emph{1st {SIG}dial Workshop on Discourse and Dialogue}, pages
  74--83, Hong Kong, China. Association for Computational Linguistics.

\bibitem[{Reimers and Gurevych(2019)}]{reimers-2019-sentence-bert}
Nils Reimers and Iryna Gurevych. 2019.
\newblock \href {http://arxiv.org/abs/1908.10084} {Sentence-bert: Sentence
  embeddings using siamese bert-networks}.
\newblock In \emph{Proceedings of the 2019 Conference on Empirical Methods in
  Natural Language Processing}. Association for Computational Linguistics.

\bibitem[{Schuster et~al.(2021)Schuster, Fisch, and
  Barzilay}]{schuster-etal-2021-get-vitac}
Tal Schuster, Adam Fisch, and Regina Barzilay. 2021.
\newblock \href {https://doi.org/10.18653/v1/2021.naacl-main.52} {Get your
  vitamin {C}! robust fact verification with contrastive evidence}.
\newblock In \emph{Proceedings of the 2021 Conference of the North American
  Chapter of the Association for Computational Linguistics: Human Language
  Technologies}, pages 624--643, Online. Association for Computational
  Linguistics.

\bibitem[{Song et~al.(2020)Song, Tan, Qin, Lu, and Liu}]{song2020mpnet}
Kaitao Song, Xu~Tan, Tao Qin, Jianfeng Lu, and Tie-Yan Liu. 2020.
\newblock \href {http://arxiv.org/abs/2004.09297} {Mpnet: Masked and permuted
  pre-training for language understanding}.

\bibitem[{Wan and Bansal(2022)}]{wan-bansal-2022-factpegasus}
David Wan and Mohit Bansal. 2022.
\newblock \href {https://doi.org/10.18653/v1/2022.naacl-main.74}
  {{F}act{PEGASUS}: Factuality-aware pre-training and fine-tuning for
  abstractive summarization}.
\newblock In \emph{Proceedings of the 2022 Conference of the North American
  Chapter of the Association for Computational Linguistics: Human Language
  Technologies}, pages 1010--1028, Seattle, United States. Association for
  Computational Linguistics.

\bibitem[{Wang and Ling(2016)}]{rt_v1_wang-ling:2016:N16-1}
Lu~Wang and Wang Ling. 2016.
\newblock \href {http://www.aclweb.org/anthology/N16-1007} {Neural
  network-based abstract generation for opinions and arguments}.
\newblock In \emph{Proceedings of the 2016 Conference of the North American
  Chapter of the Association for Computational Linguistics: Human Language
  Technologies}, pages 47--57, San Diego, California. Association for
  Computational Linguistics.

\bibitem[{Wang et~al.(2016)Wang, Nishino, Hirao, Sudoh, and
  Nagata}]{wang2016exploring_coherence}
Xun Wang, Masaaki Nishino, Tsutomu Hirao, Katsuhito Sudoh, and Masaaki Nagata.
  2016.
\newblock Exploring text links for coherent multi-document summarization.
\newblock In \emph{Proceedings of COLING 2016, the 26th International
  Conference on Computational Linguistics: Technical Papers}, pages 213--223.

\bibitem[{Wolhandler et~al.(2022)Wolhandler, Cattan, Ernst, and
  Dagan}]{wolhandler2022multi_how_multi_is_mds}
Ruben Wolhandler, Arie Cattan, Ori Ernst, and Ido Dagan. 2022.
\newblock \href {http://arxiv.org/abs/2210.12688} {How "multi" is
  multi-document summarization?}

\bibitem[{Xiao et~al.(2022)Xiao, Beltagy, Carenini, and Cohan}]{primera}
Wen Xiao, Iz~Beltagy, Giuseppe Carenini, and Arman Cohan. 2022.
\newblock \href {https://doi.org/10.18653/v1/2022.acl-long.360} {{PRIMERA}:
  Pyramid-based masked sentence pre-training for multi-document summarization}.
\newblock In \emph{Proceedings of the 60th Annual Meeting of the Association
  for Computational Linguistics (Volume 1: Long Papers)}, pages 5245--5263,
  Dublin, Ireland. Association for Computational Linguistics.

\bibitem[{Zhang et~al.(2019{\natexlab{a}})Zhang, Zhao, Saleh, and
  Liu}]{zhang2019pegasus}
Jingqing Zhang, Yao Zhao, Mohammad Saleh, and Peter~J. Liu. 2019{\natexlab{a}}.
\newblock \href {http://arxiv.org/abs/1912.08777} {Pegasus: Pre-training with
  extracted gap-sentences for abstractive summarization}.

\bibitem[{Zhang et~al.(2019{\natexlab{b}})Zhang, Kishore, Wu, Weinberger, and
  Artzi}]{zhang2019bertscore}
Tianyi Zhang, Varsha Kishore, Felix Wu, Kilian~Q Weinberger, and Yoav Artzi.
  2019{\natexlab{b}}.
\newblock Bertscore: Evaluating text generation with bert.
\newblock \emph{arXiv preprint arXiv:1904.09675}.

\bibitem[{Zhao et~al.(2022)Zhao, Strube, and Eger}]{discoscore}
Wei Zhao, Michael Strube, and Steffen Eger. 2022.
\newblock \href {http://arxiv.org/abs/2201.11176} {Discoscore: Evaluating text
  generation with bert and discourse coherence}.

\end{thebibliography}
